\documentclass[times,twocolumn,final,authoryear]{elsarticle}

\usepackage{prletters}
\usepackage{framed,multirow}

\usepackage{amssymb}
\usepackage{latexsym}

\usepackage{booktabs} 
\usepackage{amssymb}
\usepackage{comment}
\usepackage{amsmath}
\usepackage{algorithm}
\usepackage{algorithmicx}
\usepackage{algpseudocode}
\usepackage{mathrsfs} 
\DeclareMathOperator*{\argmax}{argmax}

\usepackage{multirow}
\usepackage{graphicx}
\usepackage{subfigure}
\usepackage{color}
\usepackage{bbm}
\usepackage[font=scriptsize]{caption}

\usepackage{url}
\usepackage{xcolor}
\definecolor{newcolor}{rgb}{.8,.349,.1}

\journal{Pattern Recognition Letters}

\begin{document}

\thispagestyle{empty}

\clearpage
\thispagestyle{empty}
\ifpreprint
  \vspace*{-1pc}
\fi

\clearpage

\ifpreprint
  \setcounter{page}{1}
\else
  \setcounter{page}{1}
\fi

\begin{frontmatter}

\title{Opinion Fraud Detection via Neural Autoencoder Decision Forest}

\author[1]{Manqing \snm{Dong}\corref{cor1}}
\cortext[cor1]{Corresponding author:
  }
\ead{dongmanqing@gmail.com}
\author[1]{Lina \snm{Yao}}
\author[1]{Xianzhi \snm{Wang}}
\author[1]{Boualem \snm{Benatallah}}
\author[1]{Chaoran \snm{Huang}}
\author[1]{Xiaodong \snm{Ning}}

\address[1]{School of Computer Science and Engineering, University of New South Wales, Sydney 2052, Australia}

\begin{abstract}
Online reviews play an important role in influencing buyers' daily purchase decisions.
However, fake and meaningless reviews, which cannot reflect users' genuine purchase experience and opinions, widely exist on the Web and pose great challenges for users to make right choices. Therefore, it is desirable to build a fair model that evaluates the quality of products by distinguishing spamming reviews.

We present an end-to-end trainable unified model to leverage the appealing properties from Autoencoder and random forest. A stochastic decision tree model is implemented to guide the global parameter learning process. Extensive experiments were conducted on a large Amazon review dataset. The proposed model consistently outperforms a series of compared methods.

\end{abstract}


\end{frontmatter}


\section{Introduction}
As a result of the popularity of the e-commerce and experience sharing websites, online reviews influence increasingly on people' purchase decisions. More and more people are referring to others' opinions before buying something. In view of the significance of online reviews to the success of e-commerce, many sellers intentionally post fake reviews, either positive to promote the sales of their own products or negative ones to pull down the sales of other sellers' products.
In fact, nowadays, online fraud, including the posting of fake reviews, has become a common phenomenon driven by profits \citep{wu2015spammers}. In the worst cases, some even hire a group of people for fraud.

Under such circumstances, it is paramountly important
to evaluate products based on those credible reviews rather than the suspicious (or fake) ones, to help online users make wise purchase decisions.
Until now, there have been a large body of research efforts on detecting fake or spam reviews\citep{sandulescu2015detecting,mukherjee2012spotting}, as well as spammers (i.e., the person who provides spam reviews) and spammer groups\citep{mukherjee2012spotting}. Most existing approaches achieve these purposes by extracting features from the review text, ratings, product meta-data (category and price), or review feedback (helpful/unhelpful votes etc.)\citep{jindal2008opinion}. Some researchers improve performance by considering user behavior patterns (e.g, user's interactions with products or other users) \citep{wu2015spammers} or the probabilistic distribution of users' behaviors(spamming and non-spamming) \citep{li2017bimodal}. \cite{rayana2015collective} design a collective unified framework to exploit linguistic clues of deception, behavioral footprints, or relational ties between agents in a review system.

As for the learning techniques, several previous works mainly use traditional classification methods such as Support Vector Machine (SVM) \citep{ott2011finding}, Naive Bayes Classifiers, and logistic regression\citep{jindal2008opinion}. Some researchers consider to use text analysis, for example, learning the written patterns \citep{ren2017leveraging}. Some used probabilistic methods such as unsupervised Bayesian approach \citep{mukherjee2013spotting}, and Hidden Markov models \citep{li2017bimodal}. And limited works considered using deep learning based methods \citep{lecun2015deep} for detecting the fake reviews \citep{wang2016semi}.

Deep learning methods, such as Convolutional Neural Networks (CNN) and Recurrent Neural Networks (RNN) \citep{lecun2015deep}, have achieved a great amount of success in works such as Image Recognition\citep{he2016deep}. They show advantages in dealing with high dimensional data and utilize the non-linear combination of input features -- where for detecting the fake news, we are also dealing with various features and try to find their link to the signal of fake.  Some related works includes Ma's attempt\citep{ma2016detecting}, they used RNN for detecting rumors in microblogs; and in Baohua's work\citep{wang2016semi}, they designed a semi-supervised recursive autoencoders for detecting review spam in microblogs. However, is there a way that could combine the deep learning methods and traditional classification methods, to fully extend the advantages of different methods?

In this work, we present an end-to-end trainable unified model to leverage the appealing properties of Autoencoder\citep{lange2010deep} and random forest\citep{liaw2002classification}. And the reason for combining these two methods is shown in the following. Autoencoder has been proved to be a robust algorithm which can produce unsupervised representations in feature patterns, and such representations are often more concise. Random forest is the ensemble of several decision trees, which can help prevent over fitting in each tree and shows good performance in practice. In our unified model, we use autoencoder to generate the hidden representations of the features, and take them as the input of random forest. The entire model is trained jointly via the stochastic and differentiable decision tree model, and the decision forest generates the final prediction. To summarize, we make the following contributions to the field of opinion fraud detection.

\begin{itemize}
\item We employ statistical analysis to define a list of quality measures, which utilize multiple metrics such as reviewer behavioral patterns, and review content analysis. These measures will be used as discriminative indicators of spamicity.

\item We propose to detect spam reviews with the learned quality features to infer the salient correlations between reviews and heterogeneous cross-domain historical user information. We propose a joint model by fusing autoencoder and random forest to harvest the merits in an end-to-end trainable way. The model is learned by back propagation using stochastic and differentiable decision tree model. As such, the global optimization of model parameters like splits, leafs and nodes can be obtained.

\item Experimental results on the real dataset demonstrate that the proposed model achieves the best prediction accuracy compared to several baselines and state-of-art methods.  We have made the related code and dataset open-sourced for reuse.
\end{itemize}


\section{Related Work}
Opinion spam detection has been an active field of research in recent years, which covers broad topics such as detecting singleton fake reviewers, fake reviews, or even spam groups. For example, Jindal et al. \citep{jindal2008opinion} built a logistic regression classifier with review feedback features, content characteristics, and rating related features to detect fake reviewers;  \citep{feng2012syntactic} investigated syntactic stylometry for deceptive reviews detection by using SVR classifiers; \citep{mukherjee2012spotting} propose a model for group opinion spam spotting with assistance of human experts.

Most detection methods are learning discriminate features from the linguistic \citep{ott2011finding,feng2012syntactic}, relational \citep{akoglu2013opinion}, and behavioral aspects \citep{feng2012syntactic,mukherjee2012spotting} based on review text, ratings, product meta-data (category and price), and review feedback (helpful votes), reviewer behavioral patterns and linkage structures. Original fake review detection methods mainly consider only the review text information, for example, Ott et al.\citep{ott2011finding} consider text categorization and the sentiment of the text. And later on, a lot of works combine those different aspects of features for improvements. In Rayana's work \citep{rayana2015collective}, they combine features such as review texts, time-stamps, user behavioral information and the review network. However, limited work found have conducted quality feature analysis (e.g. the importance of a feature) for evaluating the selected features.

The techniques used in detecting the opinion fraud vary a lot. Some researchers use classical classification methods such as Support Vector Machine (SVM) \cite{ott2011finding}, Naive Bayes Classifiers, and logistic regression\citep{jindal2008opinion}. Such as Vlad et al.\citep{sandulescu2015detecting} employ bag-of-words and bag-of-opinions Latent Dirichlet Allocation (LDA) model to detect singleton review spammers based on semantic similarity, in which they use cosine similarity as a measure, and topics and extracted from the reviews text, with parts-of-speech (POS) as patterns. And others may prefer statistical methods: Mukherjee et al. \citep{mukherjee2013spotting} use unsupervised Bayesian approach and considered user's behavior features and bigrams; A most recent work \citep{li2017bimodal} models the distribution of reviewers' posting rates and utilizes a coupled hidden Markov model to capture both reviewers' posting behaviors and co-bursting signals. There are also some trials on deep learning, which achieves great success in several areas in recent years. Several attempts include Wang's work\citep{wang2017liar}, which proposes using a hybrid convolutional neural network to detect fake news. Baohua\citep{wang2016semi} et al. designed a semi-supervised recursive autoencoders for detecting review spam in Weibo, a Chinese alternative to Twitter. Still, to the best of our knowledge, very limited work on hybrid deep learning models are proposed for fake opinion detection.

In this work, we propose to integrate neural networks with neural random forest, which is a differentiable model that could be fused into deep learning methods. And there have been already several attempts on leveraging the benefits of deep learning methods and random forest. One of the most important work in this area is the studies that cast the neural random forest as the structure of neural networks. The first few attempts started in the 1990s when Sethi\cite{sethi1990entropy} proposed entropy net, which encoded decision trees into neural networks. Welbl et. al \cite{welbl2014casting} further proposed that any decision trees can be represented as two-layer Convolutional Neural Networks (CNN). Since the classical random forest cannot conduct back-propagation, other researchers tried to design differentiable decision tree methods, which can be traced from Montillo's work \cite{montillo2013entanglement}. They investigated the use of sigmoidal functions for the task of differentiable information gain maximization. In Peter et al.'s work\cite{kontschieder2015deep}, they introduce a stochastic, differentiable, back propagation compatible version of decision trees, guiding the representation learning in lower layers of deep convolutional networks. Different from their work, we develop a hybrid model by combining neural decision forest with lightweight autoencoder to detect fraud opinion.

\section{Quality Feature Representation}
For detecting the fake reviews, the first thing is defining the features. In this paper, we mainly considered two scopes of features, one is user's behavioral features and one is user's review content features. The selected numerical features and the description of those features are listed in Table \ref{tab:feature}. And the categorical features will be shown later.

\begin{table}[ht!]
\caption{Feature Explanation}
\centering
\label{tab:feature}
\small
\begin{tabular}{p{250pt}}
  \toprule
    \multicolumn{1}{c}{\textbf{User's behavioral data}} \\      \midrule
    \textbf{Number of reviewed products} \\
    \textbf{Length of name} \\
    \textbf{Reviewed product structure}: the products include 54 different categories, such as book and music, here indicates the ratio of user's reviewed product's categories. \\  
    \textbf{Minimum score and maximum score} \\
    \textbf{Ratio of each score}: 1 to 5 \\
    \textbf{Number of each score} \\
    \textbf{Ratio of positive and negative ratings}: the proportion of high rates (score 4 and 5) and low rates (score 1 and 2) of a user\\
    \textbf{Entropy of ratings}: as a measure of skewness in ratings. The entropy of the user's rating can be explained by $-\sum^{5}_{i=1}p_{i}\log p_{i} $, where $p_{i}$ is the proportion that the user rated score i.\\
    \textbf{Total number of helpful and unhelpful votes} \\
    \textbf{Average number of helpful and unhelpful votes} \\
    \textbf{The ratio of the helpful and unhelpful votes} \\
    \textbf{Median, min, max number of helpful and unhelpful votes}.\\
    \textbf{Day gap}: the time gap between the user's first review and last review by days. \\
    \textbf{Review time entropy}: as a measure of skewness in user's review time by years, which could be denoted by formula  $-\sum^{m}_{j=1}t_{j}\log t_{j} $, where $m$ means the year gap between the user's first review and last review, and $t_{j}$ represent the proportion of the number of reviews a user generated for year $j$. \\
    \textbf{Same date indicator}: whether the user's first comment and the last comment are on the same date, 1 indicate the same date. \\
    \textbf{Active ratio}: the proportion that a user was active during $m$ years.\\ \midrule
    \multicolumn{1}{c}{\textbf{User's review data}} \\ \midrule
    \textbf{The Average review's ratings for the product} \\
    \textbf{Number of reviews} \\
    \textbf{Entropy of the scores}: as a measure of the skewness for the product's review ratings. Can be denoted by $-\sum^{5}_{i=1}s_{i}\log s_{i} $, where $s_{i}$ denotes the proportion for score $i$ in the whole reviews. \\
    \textbf{Comment time gap}: the duration of the first review and the last review by days. \\
    \textbf{Entropy of product's comment time}: as a measure of the skewness for the product's review times, $month$ stands for the duration of the reviews for the product by month, then the entropy of product's comment time can be denoted by $-\sum^{month}_{i=1}r_{i}\log r_{i} $, where $r_{i}$ means the ratio of the reviews created in month $i$.\\
    \textbf{User's rate for the product}. \\
    \textbf{User's helpful and unhelpful votes from others}. \\
    \textbf{User's comment time gap ratio}: the days gap between user's comment time and the first comment divided by whole comment time gap for the product. \\
    \textbf{User's comment time rank}: rank equal to 1 means this user is the first person giving the product review ratio. \\
    \textbf{The ratio of the user's comment time rank and the total number of comments}. \\
    \textbf{Review summary length}: the number of words of the summary text. \\
    \textbf{Review text semantic}: the semantic value for the review content, 1 stands for positive, 0 stands for neutral, -1 stands for negative.  \\ \bottomrule
\end{tabular}
\end{table}

We extensively studied the reviewer behavior patterns in this section for the good understanding of the difference between spamicity and non-spamicity.

First, we investigate the difference between spammers and non-spammers. We used Wilcoxon signed-rank test\citep{kerby2014simple} to evaluate this difference over the continuous data such as entropy of ratings and review time entropy. This is a non-parametric statistical hypothesis test used to compare two related samples, and determines whether their population means ranks differ. Also there is no need to make an assumption on data distribution while using the {\em Wilcoxon Signed-Rank Test}.

Table \ref{tab:pvalue} shows the p-value of the Wilcoxon Signed-Rank Test of three different alternative hypotheses:not equal to 0, less than 0, and greater than 0.
A small p-value, especially a p-value is less than 0.05, stands for the high possibility to accept the alternative hypothesis, which also indicates a significant difference between spammer and non-spammer. For example, from the first line from table 2, we could know that the entropy of ratings for spammers is significantly less than non-spammers, which indicates the spammers are tending to give similar rates.
The features in bold typeface are the features that show low significance in the distinguishing spammers and non-spammers.

\begin{table}[ht]
\caption{{\em p-value} of Wilcoxon Signed-Rank Test for Continuous Features}
\centering
\scriptsize
\label{tab:pvalue}
\begin{tabular}{c|ccc}
  \toprule
    Features & \multicolumn{3}{c}{\textbf{(\em p-value) with Alternative Hypothesis}} \\ \cline{2-4}
     & Not equal to 0 & Less than 0 & Greater than 0 \\ \hline
     & \multicolumn{3}{c}{User's history features } \\ \hline
    Entropy of ratings & 2.8e-05 & 1 & 1.4e-05 \\
    Number of reviewed products & 0.0492 & 0.0246 & 0.9754 \\
    Ratio of positive ratings & 0.0003 & 0.0001 & 0.9998 \\
    \textbf{Ratio of negative ratings} & 0.1007 & 0.9497 & 0.0503 \\
    Number of helpful votes & 0.0001 & 5.3e-05 & 0.9999 \\
    Sum unhelp & 4.2e-06 & 2.1e-06 & 1 \\
    Average help & 8.1e-08 & 4.1e-08 & 1 \\
    Average unhelp & 1.5e-11 & 7.3e-12 & 1 \\
    Time gap & 0.0024 & 0.0012 & 0.9988 \\
    Entropy of rating time & 0.0006 & 0.0003 & 0.9997 \\
    Active ratio & 0.0085 & 0.9958 & 0.0042 \\
    \textbf{Memo length} & 0.8915 & 0.5544 & 0.4457\\
    Mean rate & 0.0004 & 0.0002 & 0.9998 \\ \hline
     & \multicolumn{3}{c}{User's review data} \\ \hline
    Overall product score & 5.8e-09 & 2.9e-09 & 1 \\
    Number of comments & 2.2e-16 & 1 & 2.2e-16 \\
    \textbf{Number of first day's comments} & 0.4217 & 0.7892 & 0.2108 \\
    Product's comment time gap & 2.2e-16 & 2.2e-16 & 1 \\
    Comment rank & 2.2e-16 & 1 & 2.2e-16 \\
    Comment rank ratio &  2.2e-16 & 1 & 2.2e-16 \\
    User help & 2.2e-16 & 2.2e-16 & 1\\
    User unhelp &  2.2e-16  &  2.2e-16 & 1 \\
    Length of the summary & 2.2e-16 & 1 & 2.2e-16 \\
    \textbf{Length of the review text} & 0.5556 & 0.2778 & 0.7222 \\
    User comment time gap & 5.4e-05 & 2.7e-05 & 1\\
    User comment time gap ratio & 2.2e-16 & 1 & 2.2e-16 \\
     \bottomrule
\end{tabular}
\end{table}

\begin{figure*}[!tb]
\begin{minipage}{6.1cm}
\includegraphics[width=6.1cm]{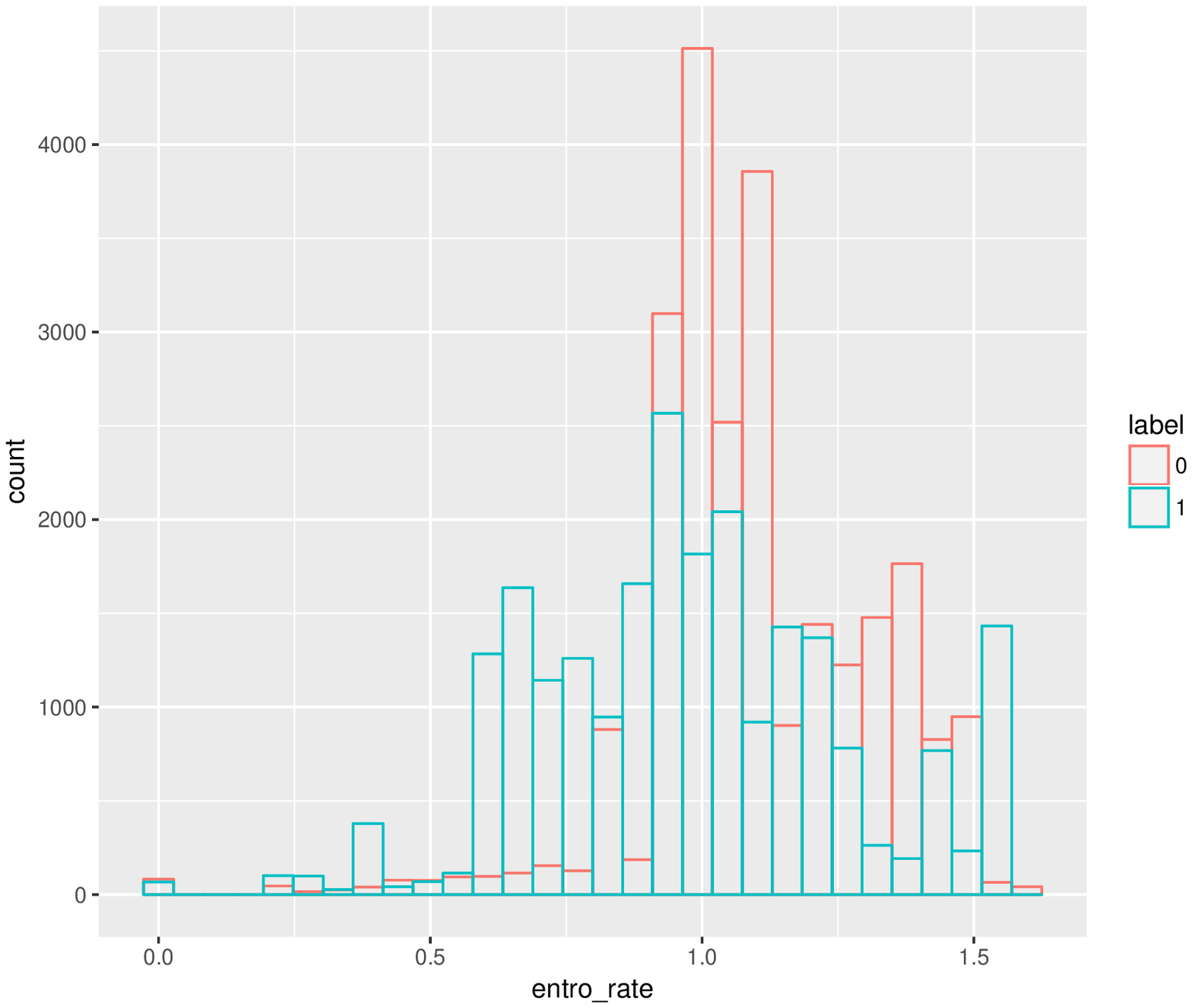}
\centering{(a)The entropy of ratings}
\end{minipage}
\begin{minipage}{6.1cm}
\includegraphics[width=6.1cm]{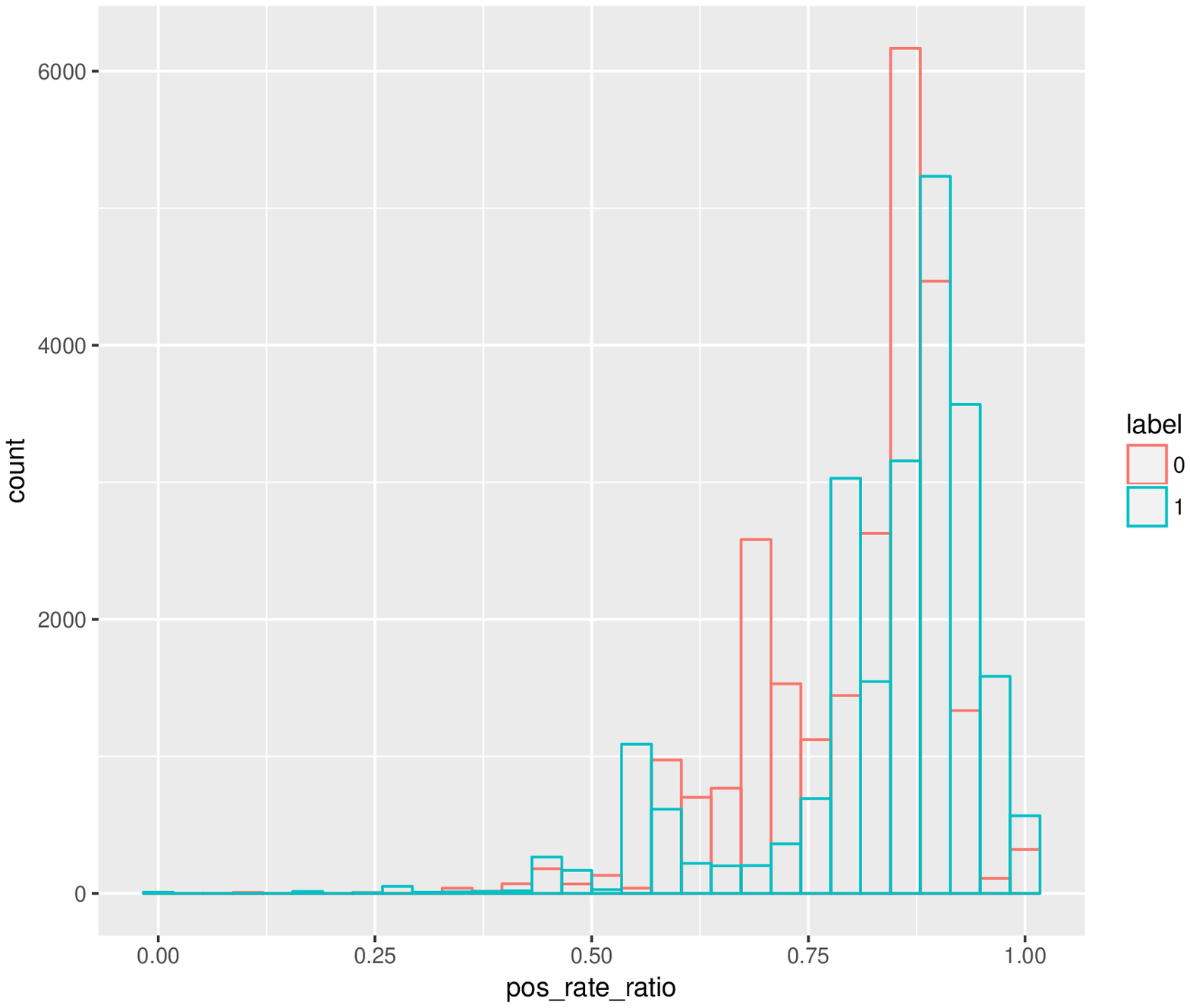}
\centering{(b)The positive rate ratio}
\end{minipage}
\begin{minipage}{6.1cm}
\includegraphics[width=6.1cm]{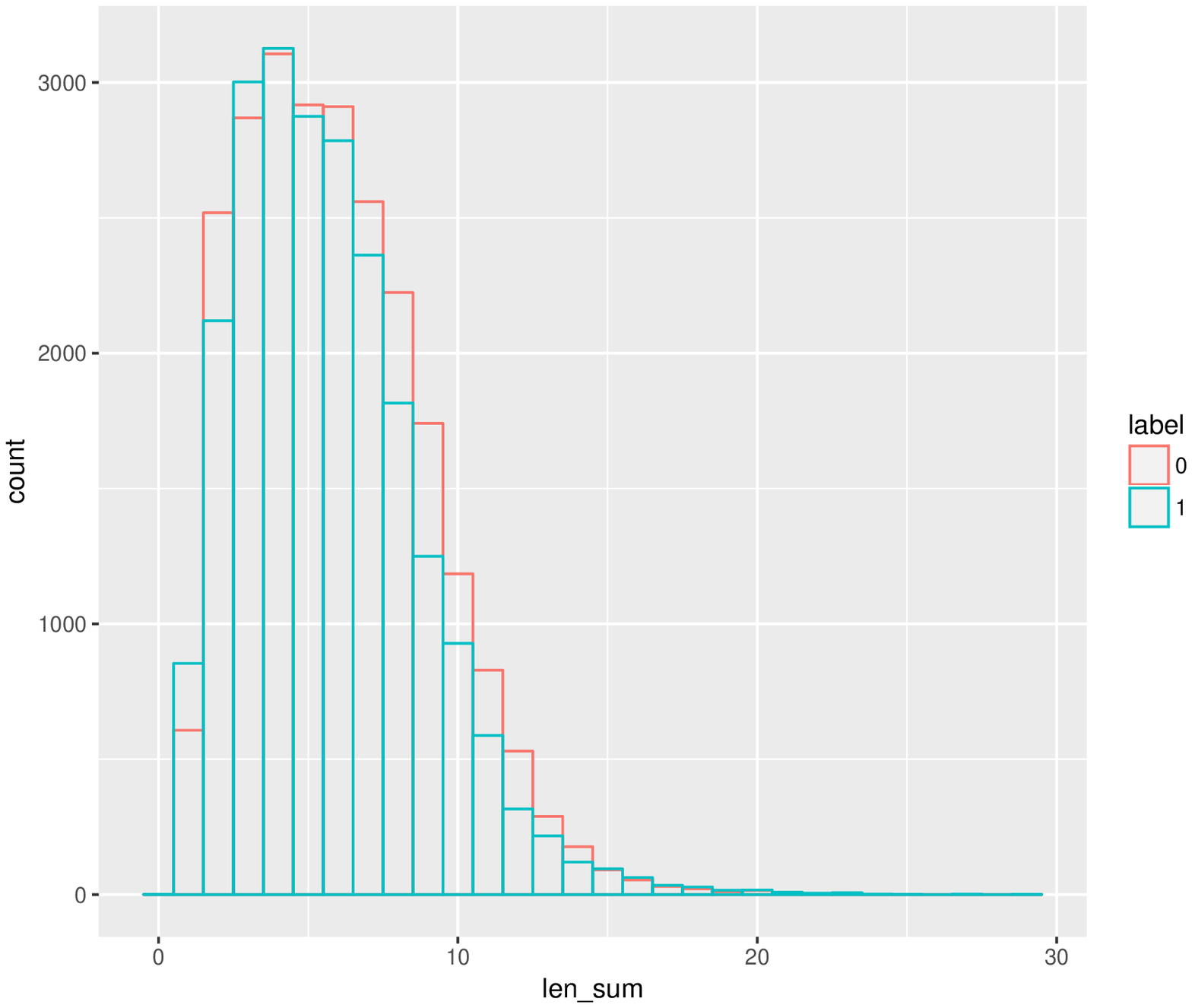}
\centering{(c)The length of the summary}
\end{minipage}
\begin{minipage}{6.1cm}
\includegraphics[width=6.1cm]{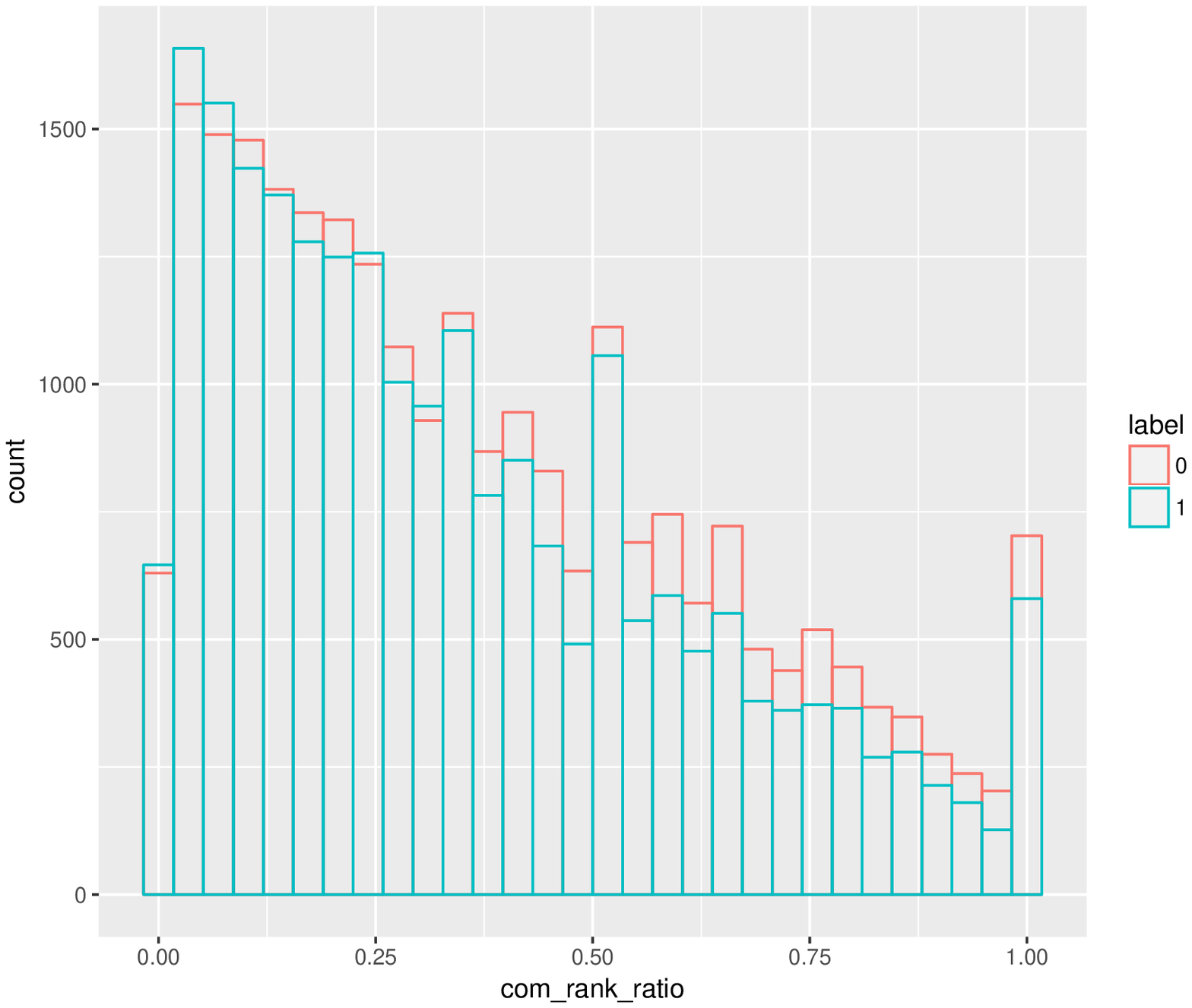}
\centering{(d)The proportion of the user's comment rank}
\end{minipage}
\begin{minipage}{6.1cm}
\includegraphics[width=6.1cm]{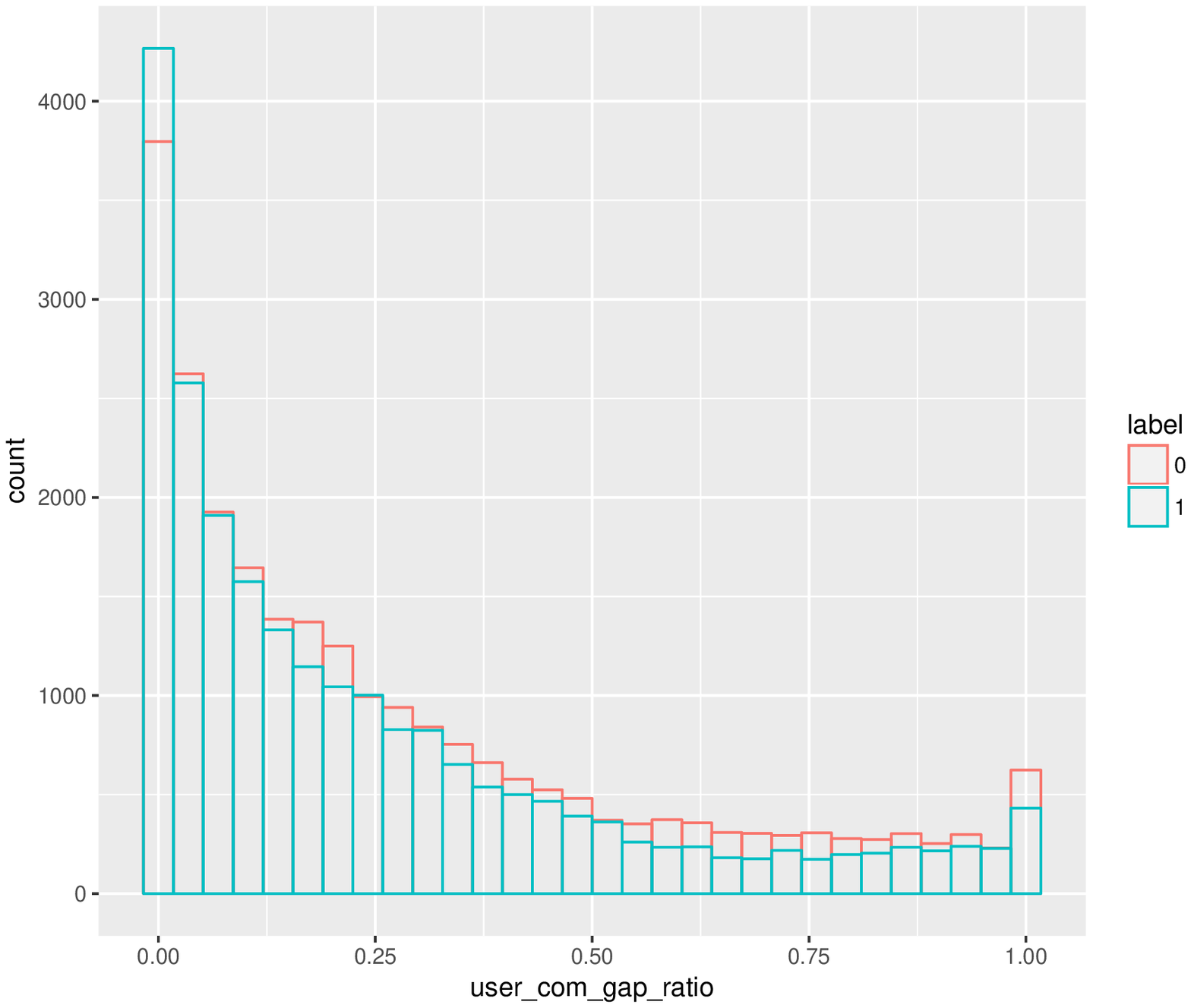}
\centering{(e)The user's time gap ratio}
\end{minipage}
\begin{minipage}{6.1cm}
\includegraphics[width=6.1cm]{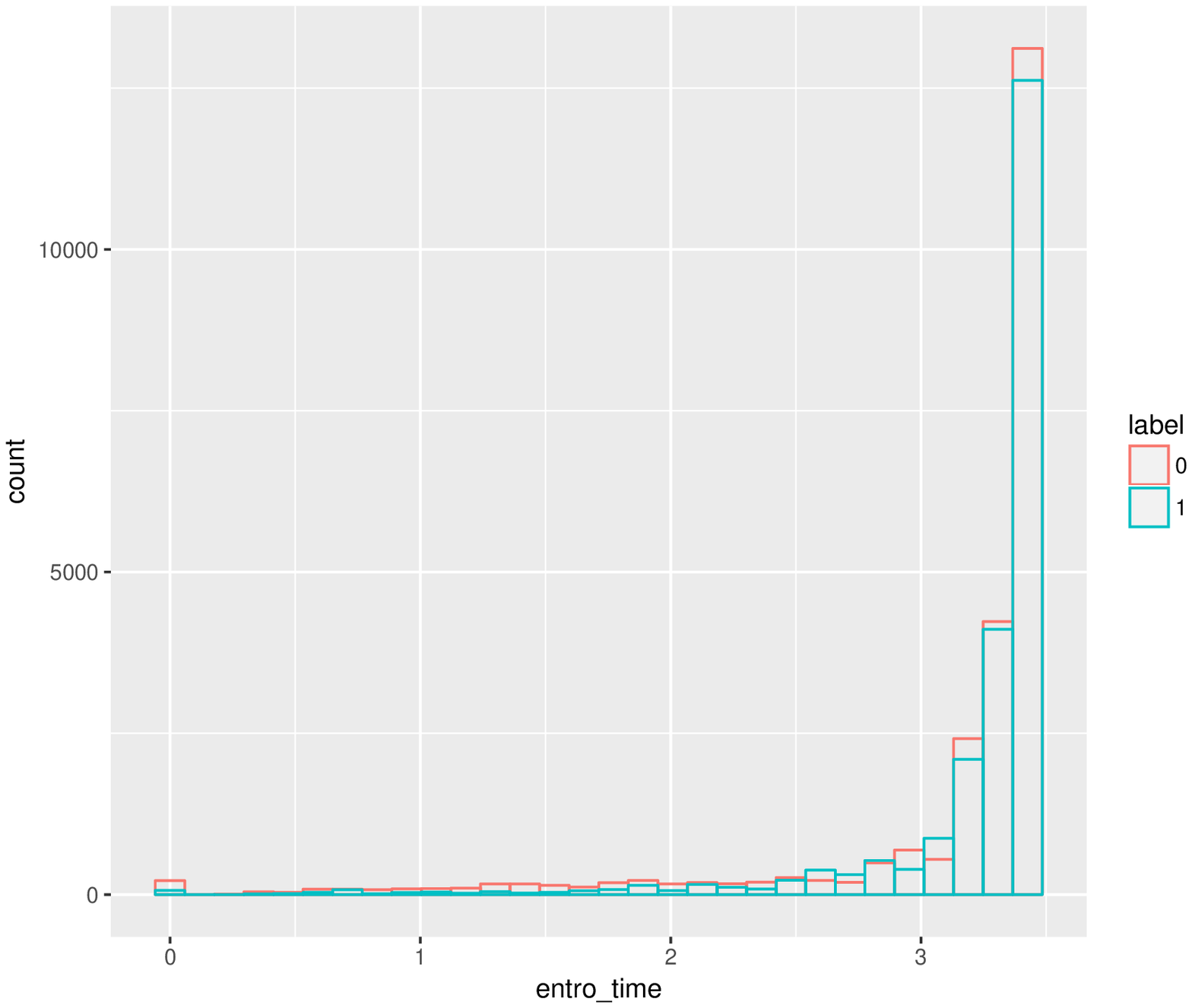}
\centering{(f)The entropy of the rating time}
\end{minipage}
\caption{The histogram of the spammer and non-spammer in (a)the entropy of the ratings, (b) the positive rate ratio, (c) the length of the summary, (d) the propotion of the user's comment rank around all reviews, (e) the user's time gap ratio for the comment time, (f) the entropy of the rating time.}
\label{fig:kernel_comparison}
\end{figure*}

Figure \ref{fig:kernel_comparison} shows a set of distinguishable patterns of review meta-data regarding density distribution. The red column stands for non-spammers and the green column stands for spammers. Figure \ref{fig:kernel_comparison}(a) shows the entropy of ratings for the spammers is lower than non-spammers, which is in consistency with our p-value showed above. Figure \ref{fig:kernel_comparison}(b) reflects that the spammers are holding the higher value than non-spammers in both low and high positive rate ratio, but lower than non-spammers in the middle. This can be explained that the spammers would likely to keep his/her rating habit, that is, a user who always rates very low scores or very high scores may likely be a spammer. Figure \ref{fig:kernel_comparison}(c) describes the number of words of the summary, we could say that spammers would like to use fewer words than non-spammers; Figure \ref{fig:kernel_comparison}(d) indicates that the spammers tend to comment early on the products; Figure \ref{fig:kernel_comparison}(e) shows that spammers would be more likely appeared in products which is released for a long time; and from Figure \ref{fig:kernel_comparison}(f) we could see spammers comment more randomly than non-spammers.

For categorical features, we apply Pearson $\chi^{2}$ test\citep{corder2014nonparametric} to evaluate the correlation between the features and the labels (whether a reviewer is a spammer or not). Pearson $\chi^{2}$ test is a statistical test applied to sets of categorical data to evaluate how likely it is that any observed difference between the sets arose by chance. The null hypothesis of the test is stating that the frequency distribution of certain events observed in a sample is consistent with a particular theoretical distribution. So if {\em p-value} is less than $0.05$ stands the feature has a high correlation with the labels.

\begin{table}[ht]
\caption{{\em p-value} of Pearson $\chi^{2}$ test for categorical features}
\centering
\label{tab:pearson}
\small
\begin{tabular}{cc}
  \toprule
  Features & p-value \\ \hline
  Min rate & 1.4e-06 \\
  \textbf{Max rate} & 0.1158 \\
  \textbf{Common name} & 0.1721 \\
  \textbf{Having memo or not} & 1 \\
  User rate & 2.2e-16 \\
  \textbf{Semantic of the summary} & 0.1554 \\
  Semantic of the review text & 2.2e-16 \\
  \bottomrule
\end{tabular}
\end{table}

In Table \ref{tab:pearson}, min/max rate is the minimal/maximal score a user had through his/her review history; common username is a binary variable -- we extracted the normal English name list from world-english.org, the weirdness shows whether user's name showed in the name list, if so is valued as 0 otherwise is 1; having memo or not stands for whether a user has self-description; semantic of the summary stands for the semantic value for the review summary content; Semantic of the summary is the text analysis over review comments. The features in bold typeface don't demonstrate the significant difference between spammers and non-spammers.

\section{Proposed Methodology}
\begin{figure}[tb!]
\centering
\includegraphics[width=0.99\linewidth]{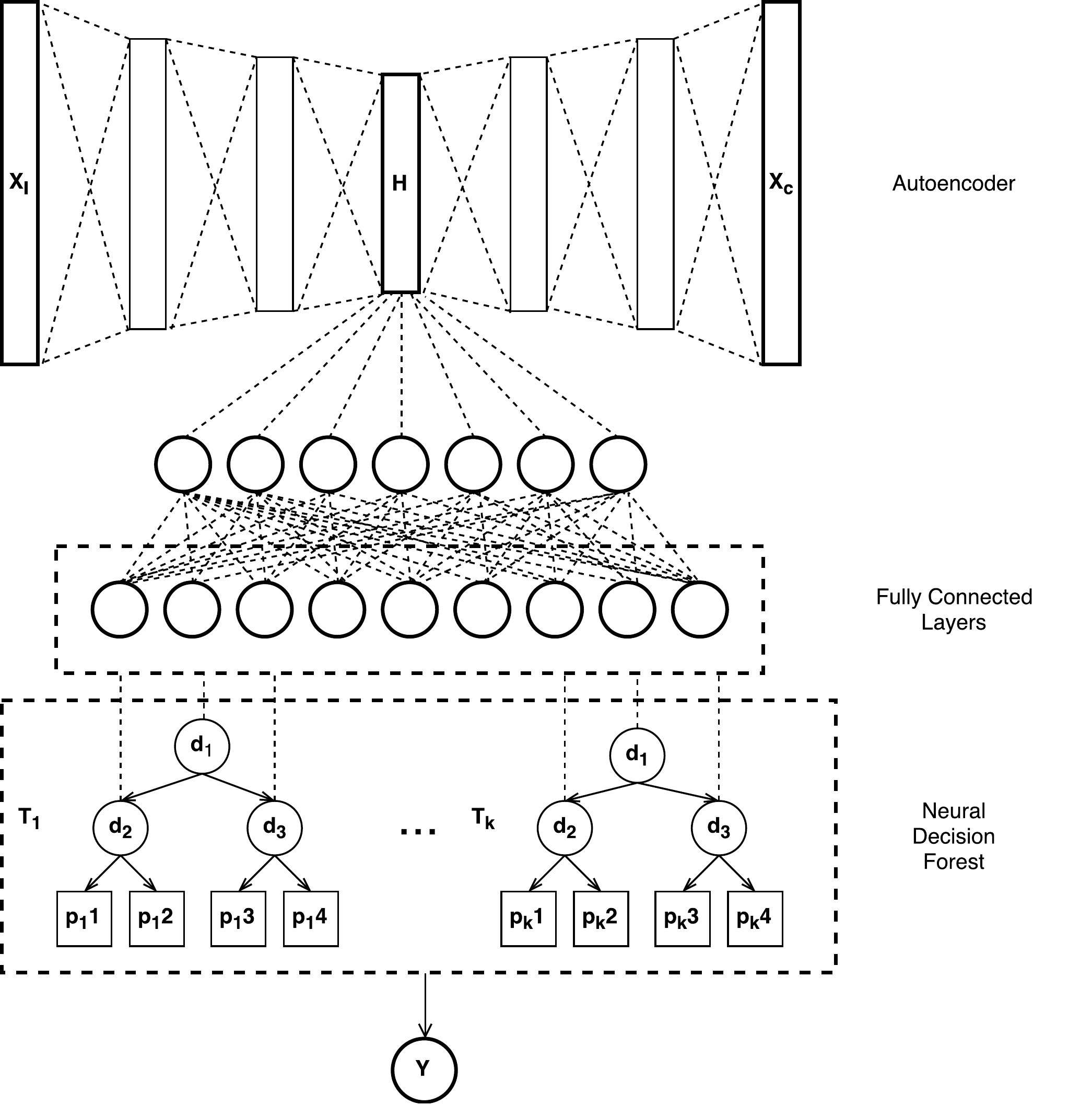}
\caption{Proposed Neural Autoencoder Decision Forest}
\label{fig:structure}
\end{figure}

In this section, we introduce our proposed methodology, which can be referred as Figure \ref{fig:structure}. The whole idea is firstly to initialize all the parameters, and feed the features into the input layer $X_{I}$ of autoencoder; also the cleaned features (the reconstruction of input layer $X_{I}$) are regarded as $X_{C}$. Then we get the hidden layer $H$, whose nodes will be followed by fully connected layers and reaches the nodes in the tree. After that, the nodes in the tree layer are delivered to the nodes in $k$ decision trees. Decision nodes will make a decision and give a prediction by averaging the decision results. Then we update all the parameters by minimizing the cost between the prediction and the real value, and iteratively optimize the whole process till to reach the optimal results.

Auto-encoder is a neural network\citep{lecun2015deep} that is trained to attempt to copy its input to its output. And it always consists of two parts, namely the encoder and the decoder. Internally, it has a hidden layer $H$ describes a code used to represent the input.

An auto-encoder takes an input $X_{I} \in \mathbb{R}^n$, where $X$ is the input feature vectors, and $n$ is the number of the features.

First we map it (with an encoder) to a hidden representation $H \in \mathbb{R}^m$,
\begin{equation}
    H = f_{E}(W_{E}X_{I}+b_{E})
\end{equation}
where $f_{E}(*)$ is a non-linearity such as the sigmoid function\citep{hecht1992theory}. And $W_{E}$ and $b_{E}$ are weights and bias variables used in encoder part

The latent representation $H$ or code is then mapped back (with a decoder) into a reconstruction $X_{C}$,
\begin{equation}
    X_{C} = f_{D}(W_{D}H+b_{D})
\end{equation}
where $X_{C}$ is seen as a prediction of $X_{I}$, given the code $H$. And $W_{D}$and $b_{D}$ are weights and bias variables used in decoder part, they are optimized such that the average reconstruction error is minimized.

The reconstruction error can be measured in many ways, depending on the appropriate distributional assumptions on the input given the code. For example, the traditional squared error $L_{AE}(X_{I},X_{C}) = \parallel X_{I} - X_{C} \parallel^2 $ can be used.

Then let the hidden layer $H$ go through several fully connected layers, we get $X_{T}$ as the input nodes for the neural decision forest

\begin{equation}
    X_{T} = g(W_{F}H+b_{F})
\end{equation}

where $W_{F}$ and $b_{F}$ are the weights used in the fully connection layers, and similarly, the $g(*)$ is the activation function. And in the next we will introduce how input those tree nodes $X_{T}$ into the tree layer.

For an end-to-end neural network, each part should be differentiable, where traditional random forest is not suitable. Thus here in our work we utilized a differentiable version of random forest, which is called neural decision forest.

The neural decision forest\citep{kontschieder2015deep} is designed by Kontschieder et al., each node in the decision tree holds probabilistic distribution thus make the decision tree differentiable. Below explain the details.

Suppose we have number of $K$ trees, for each tree, it is a structured classifier consisting of decision(or split) nodes $\mathcal{D}$ and prediction(or leaf) nodes $\mathcal{L}$. Where the leaf nodes are the terminal nodes of the tree, and each prediction node $\ell \in \mathcal{L}$ holds a probability distribution $p_{\ell}$ over $Y$. Each decision node $d \in \mathcal{D}$ holds a decision function $D_{d}(X_{T};\Theta) \in [0,1]$, which stands for the probability that a sample reached decision node $d$ and then be sent to the left sub-tree, for each decision nodes, suppose it holds the representation:

\begin{equation}
D_{d}(X_{T};\Theta) = \sigma(f_{d}(X_{T}))
\end{equation}

where $\sigma(x) = (1+e^{-x})^{-1}$ is the sigmoid function, and $f_{d}(X_{T})$ is the transfer function for $X_{T}$,

\begin{equation}
f_{d}(X_{T}) = W_{T}X_{T}
\end{equation}

$\Theta$ stands for the set of previous parameters $\{W_{E},W_{D},W_{F},W_{T},b_{E},b_{D},b_{F}\}$.

About constructing the tree, here we suppose all the trees are following a classical binary tree structure, which means each node has two subtrees, then after defining the depth $n\_depth$ of a tree, suppose the depth is 1, just like what shows in figure 1, then the decision nodes in depth 1 is 2 and the number of leaf nodes is 4. More generally, once define the depth $n\_depth$, the number of decision nodes would be $2^{n\_depth}$, and the number of leaf nodes would be $2^{n\_depth + 1}$.

And then the probability of a sample reach tree $k$ to be predicted as class $y$ would be:

\begin{equation}
\mathbb{P}_{T_{k}}[y|X_{T},\Theta,P] = \Sigma_{\ell \in \mathcal{L}} P_{\ell_{y}}(\Pi_{d \in \mathcal{D}}D_{d}(X_{T};\Theta)^{\mathbbm{1}_{left}}\overline{D_{d}}(X_{T};\Theta)^{ \mathbbm{1}_{right}})
\end{equation}
where $\overline{D_{d}}(h;\Theta) = 1 - D_{d}(h;\Theta) $, and $\mathbbm{1}_{left}$ indicates the indicator function for the nodes go left.

For example, in tree 1 $T_{1}$ in figure 1, to get to the leaf node $p_{1}$, here the $\Pi_{d \in \mathcal{D}}D_{d}(h;\Theta)^{\mathbbm{1}_{left}}\overline{D_{d}}(h;\Theta)^{ \mathbbm{1}_{right}}$ equals to $d_{1}d_{2}$, which means the probability that a sample reach leaf node $p_{1}$, and for $p_{2}$, this probability will be $d_{1}\overline{d_{2}}$. and $P_{\ell_{y}}$ means the probability for the nodes in leaf $\ell$ predicted to be label $y$. Here, in our problem, the $P_{\ell_{y}}$ will have the form $P_{\ell_{y}}=(p_{0},p_{1})$, where $p_{0}$ stands for the probability of the leaf node to be $0$ non spammer.

Then for the forest of decision trees, it is an ensemble of decision trees $\mathcal{F} = \{ T_{1},...,T_{K} \}$, which delivers a prediction for sample $x$ by averaging the output of each tree, which can be showed from:

\begin{equation}
\mathbb{P}_{\mathcal{F}}[y|x] = \frac{1}{K}\Sigma_{k=1}^{K}\mathbb{P}_{T_{k}}[y|x]
\end{equation}

Then the prediction for the label y would be:
\begin{equation}
\hat{y} = \argmax_{y}\mathbb{P}_{\mathcal{F}}[y|x]
\end{equation}

The loss for the neural decision forest is the average of the loss of each tree. And in the training process, the total loss is the average of all the training samples.

So the whole loss functions can be defined as:

\begin{equation}
L_{F}(x,y,\Theta,P) = E_{x\in X}(E_{T \in \mathcal{F}}(L_{T}(x,y,\Theta,P)))
\end{equation}

And here, for we are dealing with a classification problem, the loss function we chose here is defined as:
\begin{equation}
\begin{split}
L_{T}(x,y,\Theta,P) =& -(y\times log(\mathbb{P}_{T}[y|x,\Theta,P]) \\
 & + (1-y)log(\mathbb{P}_{T}[y|x,\Theta,P]) ) \\
 =& -log(\mathbb{P}_{T}[y|x,\Theta,P])
\end{split}
\end{equation}
where y=0 or 1.

Then our final optimal function is minimizing the whole loss function which is finding:

\begin{equation}
\mathcal{L}(X_{I},y,\Theta,P) = \arg \min_{\Theta} L_{AE}(X_{I},X_{C}) + L_{F}(x,y,\Theta,P)
\end{equation}

For learning the parameters, the optimization function we chose here is the RMSProp\citep{ruder2016overview}, which is a mini-batch version of RProp, and RProp is equivalent to using the gradient but also dividing by the size of the gradient, the parameters are updating by:

\begin{equation}
g_{t} = \nabla \mathcal{L}(\Theta_{t-1}; \mathcal{B},P)
\end{equation}

\begin{equation}
G_{t} = G_{t} + g_{t} \odot g_{t}
\end{equation}

\begin{equation}
\Theta_{t} = \Theta_{t-1} - \frac{\eta}{\sqrt{G_{t}+ \epsilon}} \odot g_{t}
\end{equation}

where $g_{t}$ is the gradient, the $\eta$ and $\epsilon$ are learning rates, and $\mathcal{B}$ is a random subset of training data set.

For learning the leaf node distribution probability P, we still follow a RProp process but add a softmax function to the updated parameters.

\begin{equation} \label{eq1}
\begin{split}
g_{t}^{P} &= \nabla \mathcal{L}(P_{t-1}; \mathcal{B},\Theta)
 = \nabla L_{F}(P_{t-1};x,y,\Theta)
\end{split}
\end{equation}

\begin{equation} \label{eq2}
G_{t}^{P} = G_{t}^{P} + g_{t}^{P} \odot g_{t}^{P}
\end{equation}

\begin{equation} \label{eq3}
P_{t} = P_{t-1} - \frac{\eta^{P}}{\sqrt{G_{t}^{P}+ \epsilon}} \odot g_{t}^{P}
\end{equation}

\begin{equation}\label{eq4}
P_{t} = softmax(P_{t})
\end{equation}

The details for how to calculate the derivative of loss function for each parameters could refer from paper \cite{kontschieder2015deep}. And below is the pseudo-code for our algorithm for both training process and testing process.

\begin{algorithm}[tb]
        \caption{Training Process}
        \begin{algorithmic}[1] 
            \Require Feature vector of a review $X_{I}$, Label for the review (fake or not) $y$, number of epoch $n\_epoch$, number of trees $n\_tree$, number of depth for a tree $n\_depth$
            \State Generalizing the structure for the forest with $n\_tree$ and $n\_depth$
            \State Initialize parameters $\Theta, P$ randomly
            \State Randomly shuffle the data sets
            \For{i \textbf{in} 1:$n\_epoch$}
                \State break data sets with $n\_batch$ of $batch\_size$ piece of data sets
                \For{j \textbf{in} 1:$n\_batch$}
                    \State Update $\Theta$ with RProp optimization.
                \EndFor
                \State Update P by \eqref{eq1} to \eqref{eq4}
            \EndFor
        \end{algorithmic}
\end{algorithm}

\begin{algorithm}[tb]
        \caption{Prediction with proposed approach}
        \begin{algorithmic}[1] 
            \Require Feature vector of a review $X$
            \Ensure Label for the review (fake or not) y
            \State Get hidden representation for $X$ by an encoder:
            \State $H =  f_{E}(W_{E}X_{I}+b_{E})$
            \State Get tree input layer from fully connected layers:
            \State $X_{T} = g(W_{F}H+b_{F})$
            \For{i \textbf{in} 1:I}
                \State p(y=i) = 0
                \For{k \textbf{in} 1:K}
                    \State $p = P_{k}[y=i|X_{T}, \Theta, P]$
                    \State $p(y=i)=p(y=i)+p$
                \EndFor
                \State p(y=i) = $\frac{1}{K}p(y=i)$
            \EndFor
            \State $y = \max_{i}p(y=i)$
            \State \Return{y}
        \end{algorithmic}
\end{algorithm}

\subsection{Computational Complexity}
Our proposed process mainly contains three parts, the autoencoder part, the fully connected layers, and the neural random forest part.

For each epoch, for the autoencoder part, the time complexity is $\mathcal{O}(\Sigma_{l}^{2*n\_AE}n_{l-1}^{AE}n_{l}^{AE})$, where $n\_{AE}$ is the number of layers from input layer to hidden layer, $l$ is the layer index, $n_{l}^{AE}$ is the number of nodes in an autoencoder layer $l$.

For the fully connected layers part, the time complexity is $\mathcal{O}(\Sigma_{l}^{n\_FC}n_{l-1}^{FC}n_{l}^{FC})$, where $n\_FC$ is the number of layers for fully connected layers, and similarly, $n_{l}^{FC}$ is the number of nodes in fully connected layer $l$.

As for the neural random forest part, the time complexity is $\mathcal{O}(n_{n\_FC}\times n\_tree \times n\_leafnodes)$, where $n_{n\_FC}$ is the number of output nodes from fully connected layers, $n\_tree$ is the number of trees in a forest, and $n\_leafnodes$ is the number of leafnodes in a tree.

For the whole training process, the whole time complexity should multiply $n\_epoch$ and $n\_batch$, which are the number of epochs and number of batch of the dataset.

\section{Experiments}
We evaluate proposed approach using a real-world dataset. In the following section we first describe our dataset, and then compared existing methods with the presentation of results and analysis. Also, experiment settings and parameter tuning are explained, as well as the impact of features.

\begin{figure}[ht]
\centering
\begin{minipage}[t]{6cm}
\includegraphics[width=6cm]{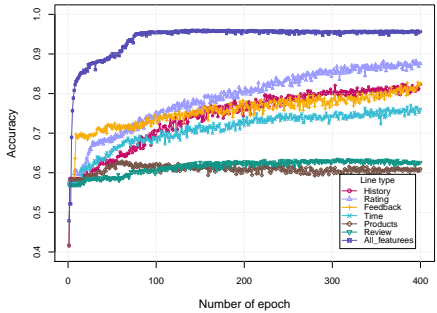}
\centering{(a)}
\end{minipage}
\begin{minipage}[t]{6cm}
\includegraphics[width=6cm]{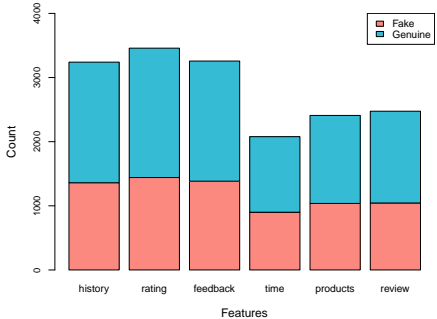}
\centering{(b)}
\end{minipage}
\caption{(a) accuracy with different feature subset, the blue line shows the accuracy using full feature set; (b) predicted results using different features. }
\label{fig:fea}
\end{figure}

\subsection{Dataset Description}
The dataset used for our work is based on the public collected Amazon review data set from \cite{he2016ups}, and ground-truth spamming review dataset from \cite{mukherjee2012spotting}. We preprocessed the dataset as follows. We obtained a manual labeled subset for 815 unique users in more than 2000 potential groups. After averaging the manually labeled spamming score (ranging from 0 and 1, more the score approaches 1, more likely this user is a spammer.), we labeled users with the average score less than 0.5 as a non-spammer and labeled them 0, otherwise, the spammers are labeled 1. Filtering is also carried out to handle missing values.
To control the high appearance of some users, who wrote even higher than 500 reviews, we set a threshold(20 reviews) for selecting user's reviews. Then we got 7950 reviews as our dataset, which is used for spamming activity detection. We regard reviews written by spammers as fake reviews and whose number counts to 3363.

\subsection{Impact of Features}
In this section, we evaluated the impact of different types of feature subsets on the performance. All features are stated in Table 1, and we divide the features into 6 scopes:
\begin{itemize}
    \item History records: number of reviewed products; length of name; reviewed product structure.
    \item Rating signal: minimum score and maximum score; Ratio of each score (1 to 5) ; number of each score; Ratio of positive and negative ratings; entropy of ratings.
    \item Feedback signal: total number of helpful and unhelpful votes; average number of helpful and unhelpful votes; the ratio of the helpful and unhelpful votes; median, min, max number of helpful and unhelpful votes.
    \item Time signal: day gap; review time entropy; same date indicator; active ratio.
    \item Product's comment information: the average review’s ratings for the product; number of reviews; entropy of the scores; comment time gap; entropy of product’s comment time.
    \item User's review information: user’s rate for the product; user’s helpful and unhelpful votes from others; user’s comment time gap ratio; user’s comment time rank; the ratio of the user’s comment time rank and the total number of comments; review summary length; review text semantic.
\end{itemize}

We run the proposed neural autoencoder decision forest using these 6 types of feature, as well as full feature set. And we draw the accuracy line chart and the proportion of the predicted results for each type of features, which is shown in figure \ref{fig:fea}.

From Figure~\ref{fig:fea}, we can observe that the performance with full feature set consistently outperforms the prediction results with single type of feature set; among the six type of features, user's rating signal is most helpful for predicting fake reviews. Which demonstrate the effectiveness of designated various types of features in this work, and the effectiveness of our proposed model for combining different scope of features to final prediction.

\subsection{Parameter Tuning}
Last section we showed the effectiveness of our model for combining the features, and in this part, we are going to tuning the parameters. We mainly considered the parameters: the number of depth of each tree in the neural decision forest; with or without fully connected layers between autoencoder and neural decision forest; number of layers in autoencoder; the normalization methods \citep{ioffe2015batch}; the batch size; and the number of trees in a forest.

In the training process, among the 7950 reviews, we randomly selected 4000 samples, and initialized the weights and biases of autoencoder with normal distributed random numbers. The default set for our model is 2 layers of autoencoder, with one fully connected layer after the autoencoder, and 5 trees with depth 2 in the neural random forest. The batch size is 100 and we take z-score as our normalization method. Then we iterated the model with 400 epochs.

\begin{figure*}[htb!]
\begin{minipage}[t]{6.15cm}
\centering
\includegraphics[width=5.6cm]{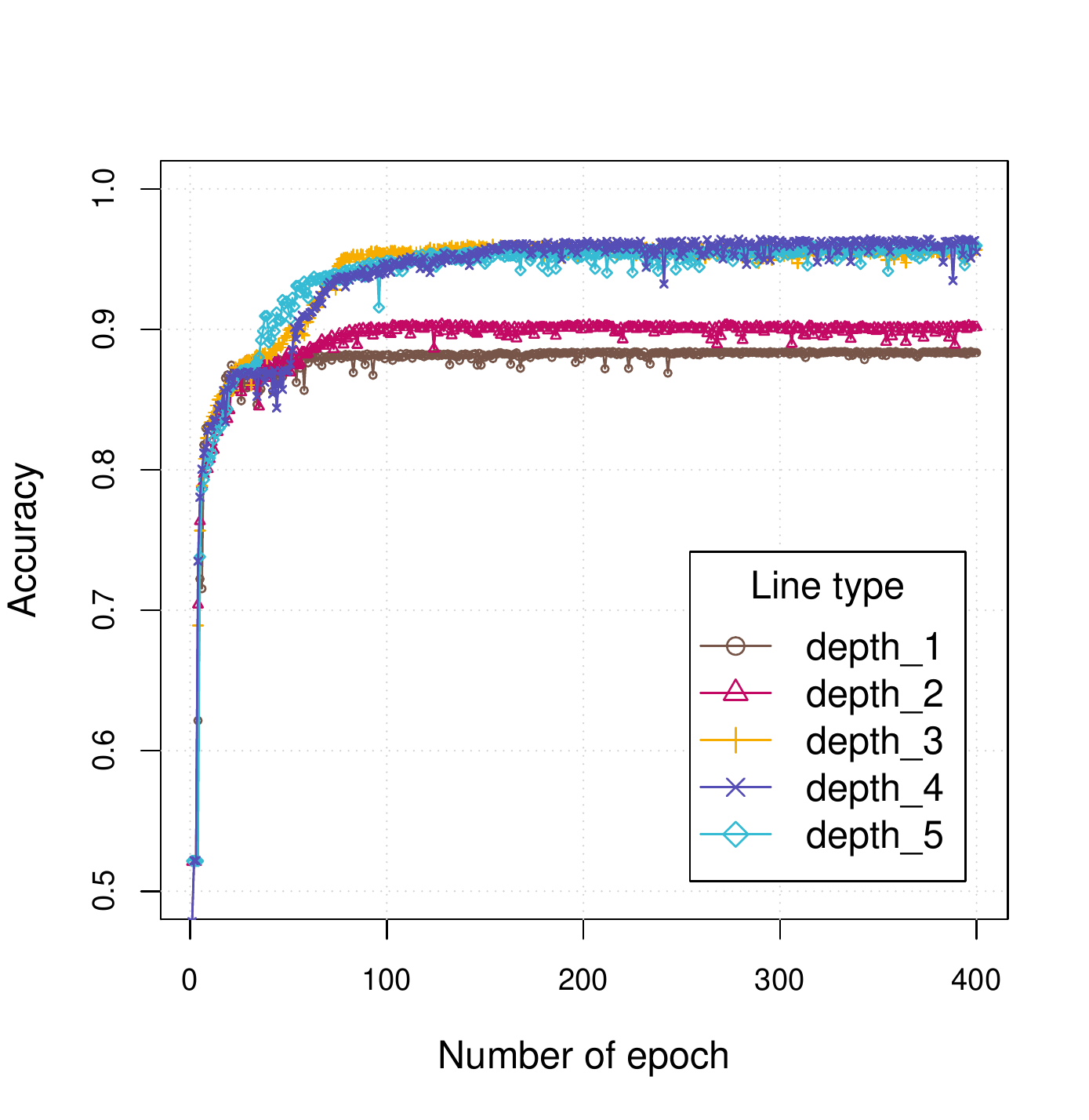}
\centering{(a)}
\end{minipage}
\begin{minipage}[t]{6.15cm}
\centering
\includegraphics[width=5.6cm]{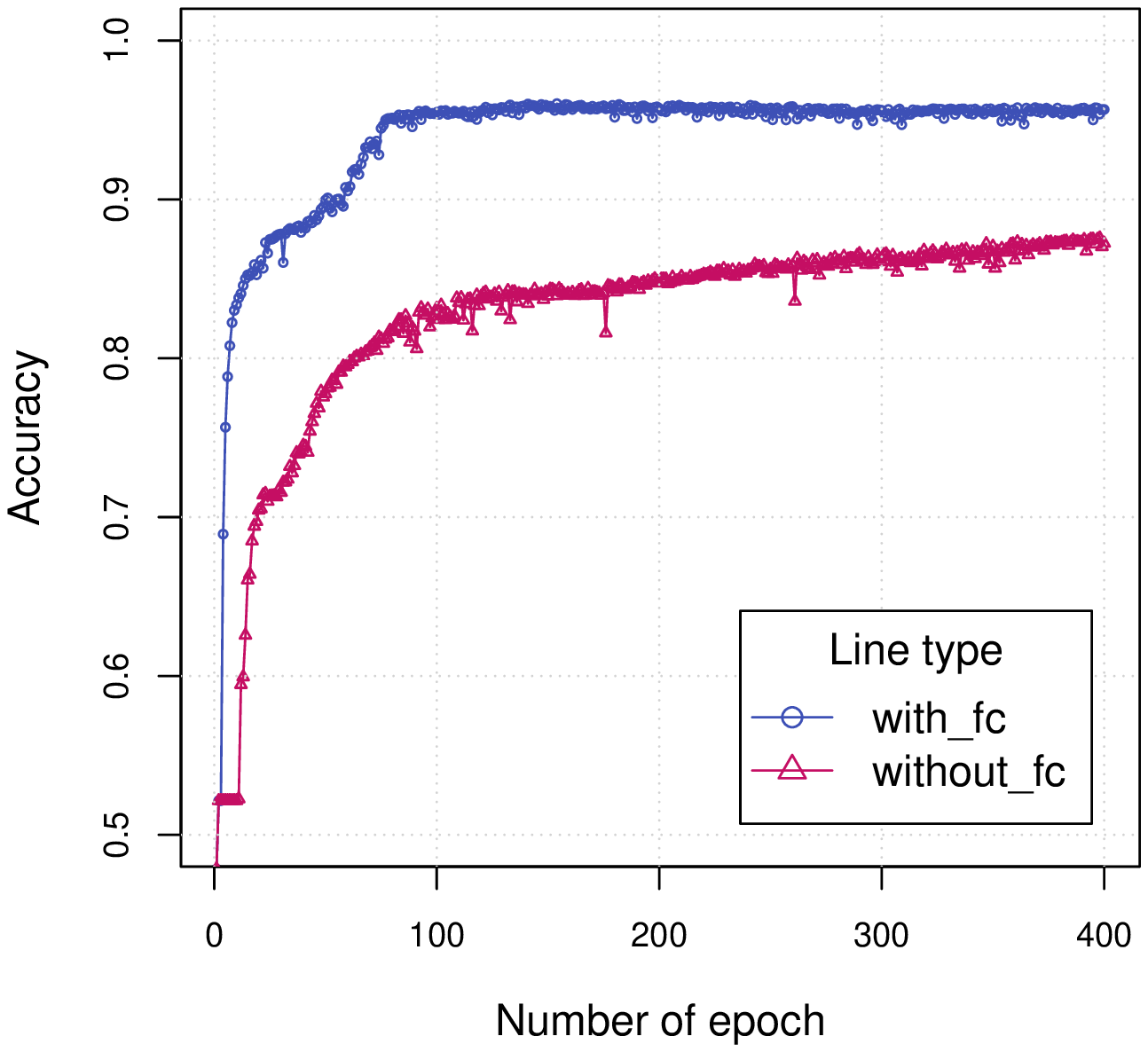}
\centering{(b)}
\end{minipage}
\begin{minipage}[t]{6.15cm}
\centering
\includegraphics[width=5.6cm]{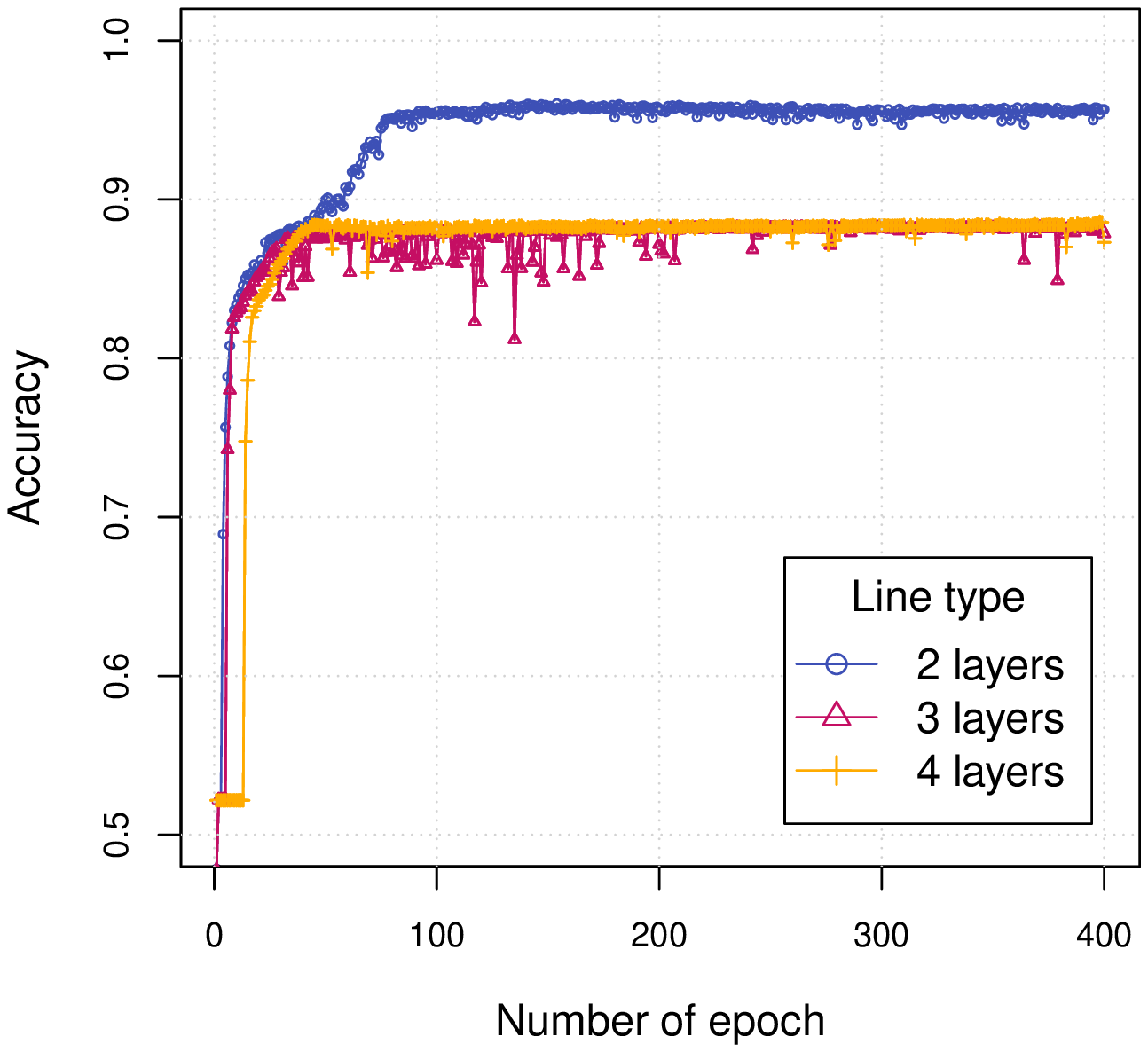}
\centering{(c)}
\end{minipage}
\begin{minipage}[t]{6.15cm}
\centering
\includegraphics[width=5.6cm]{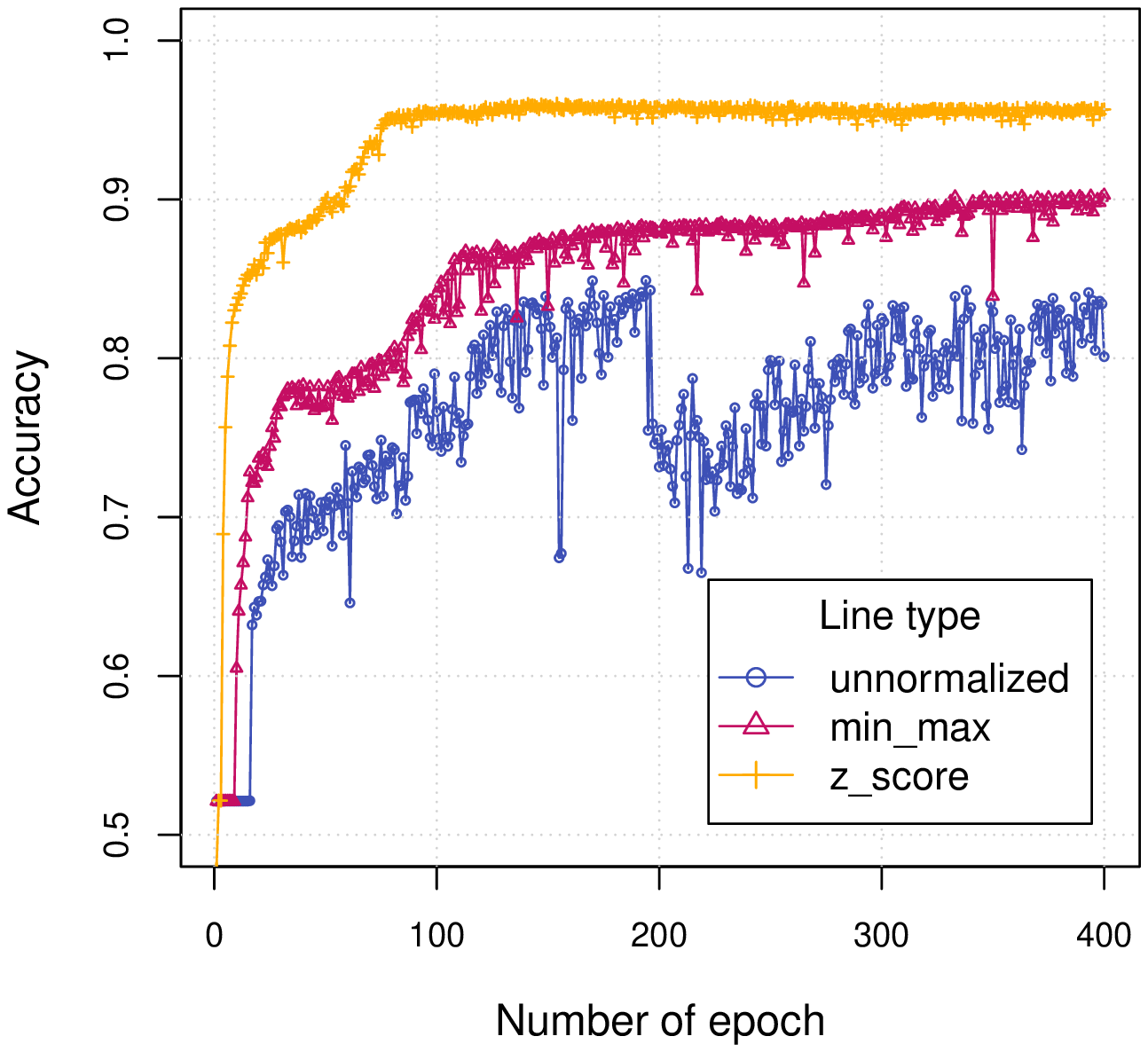}
\centering{(d)}
\end{minipage}
\begin{minipage}[t]{6.15cm}
\centering
\includegraphics[width=5.6cm]{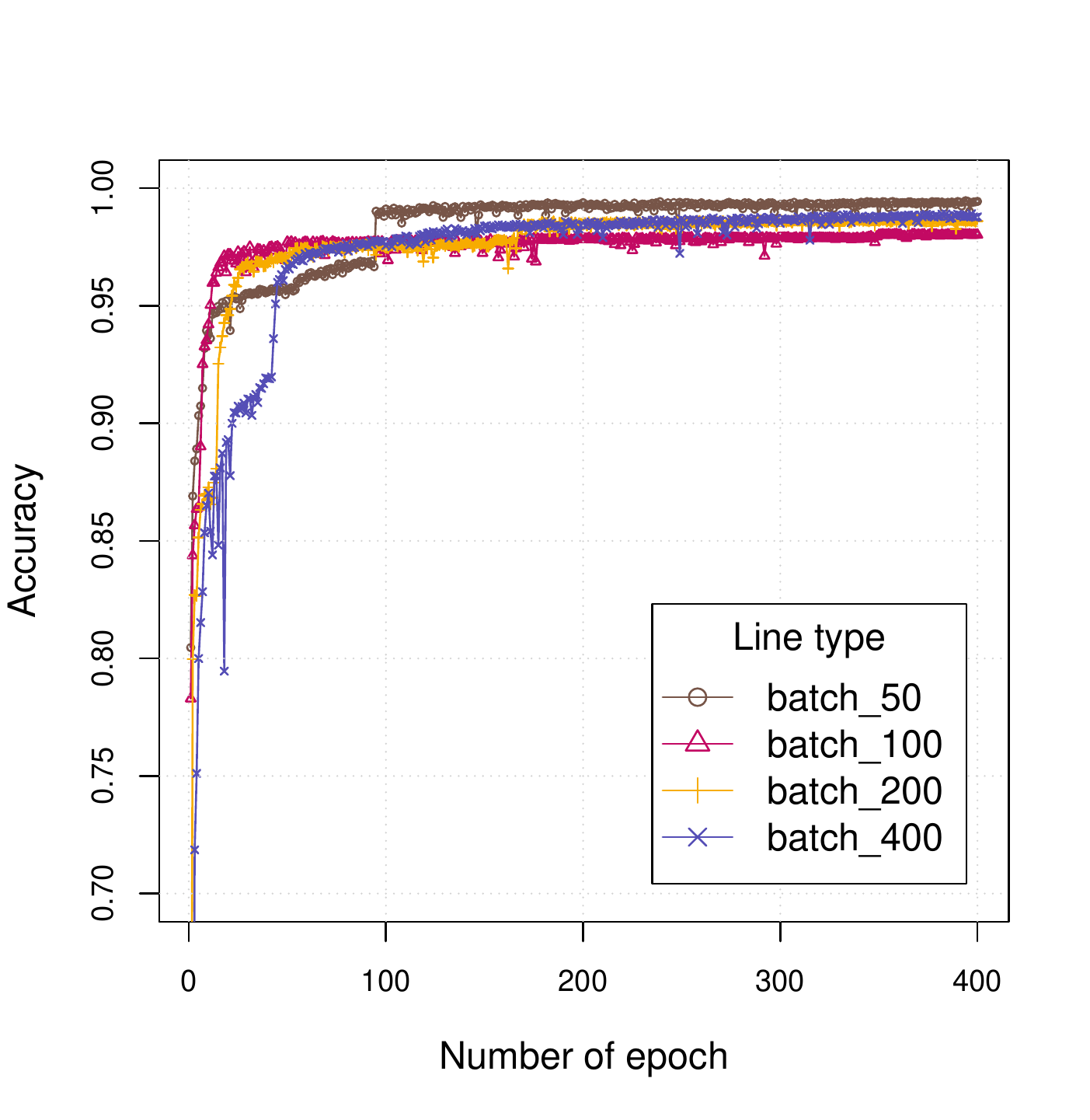}
\centering{(e)}
\end{minipage}
\begin{minipage}[t]{6.15cm}
\centering
\includegraphics[width=5.6cm]{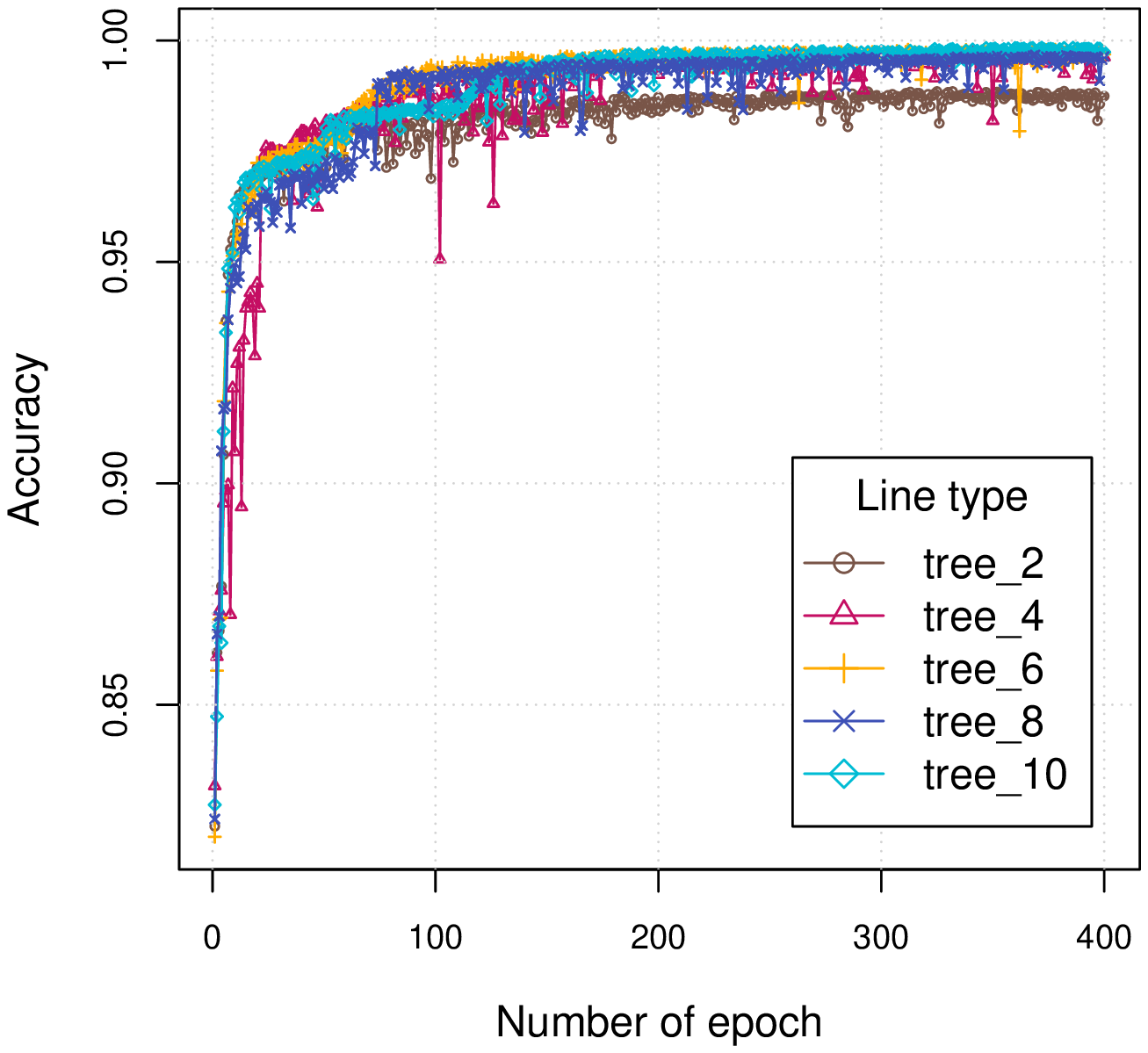}
\centering{(f)}
\end{minipage}
\begin{minipage}[t]{6.15cm}
\centering
\includegraphics[width=5.6cm]{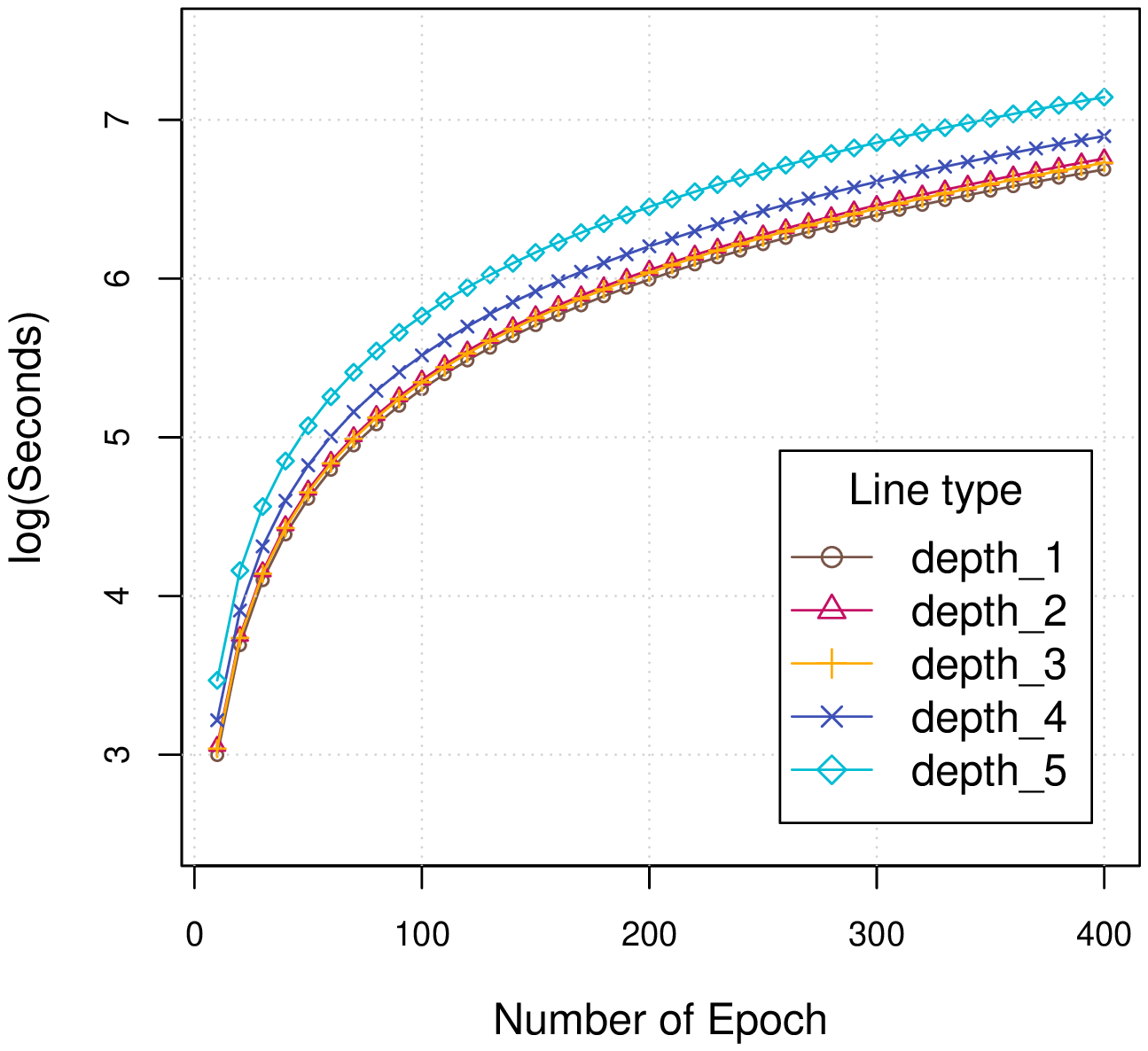}
\centering{(g)}
\end{minipage}
\begin{minipage}[t]{6.15cm}
\centering
\includegraphics[width=5.6cm]{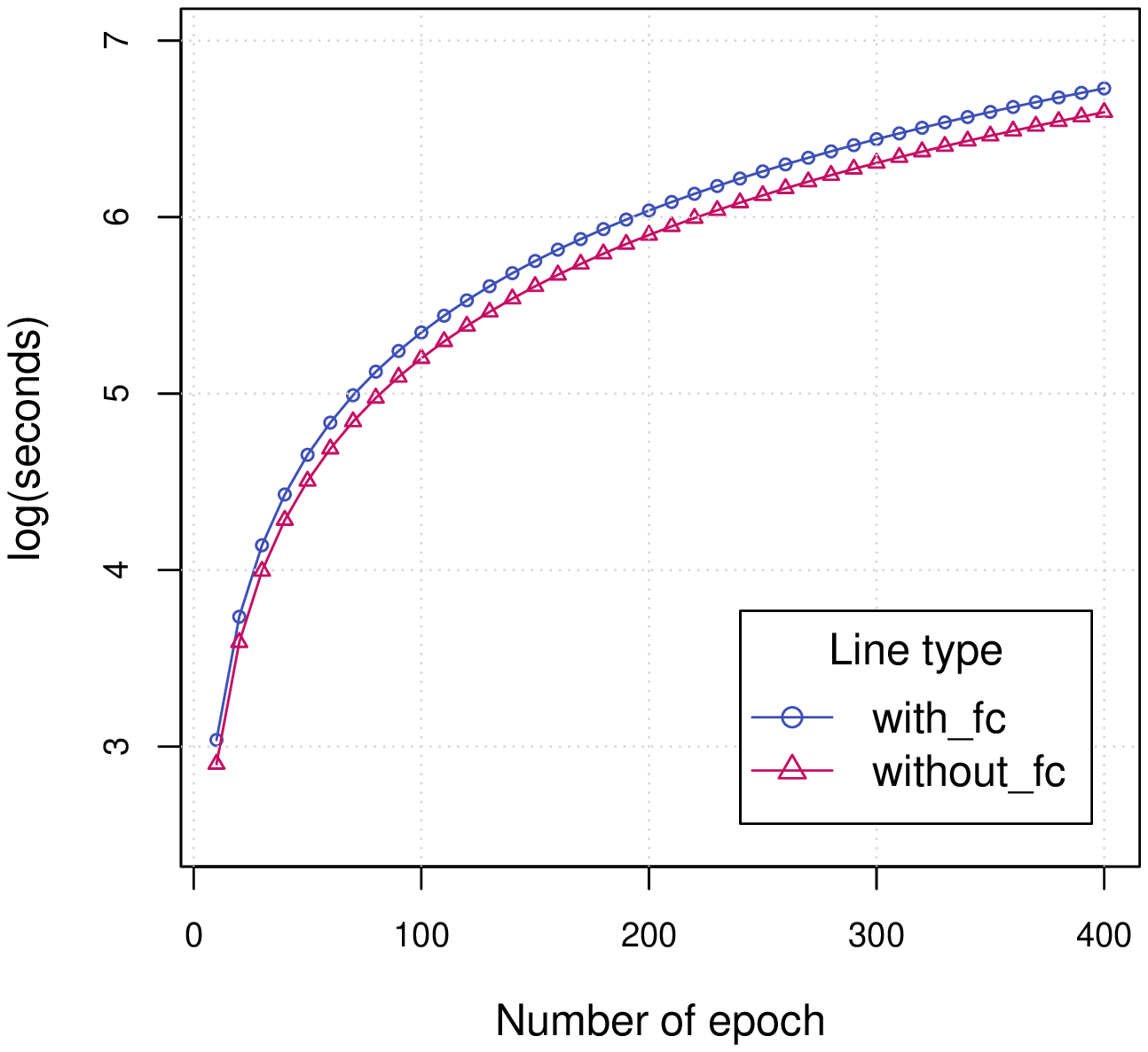}
\centering{(h)}
\end{minipage}
\begin{minipage}[t]{6.15cm}
\centering
\includegraphics[width=5.6cm]{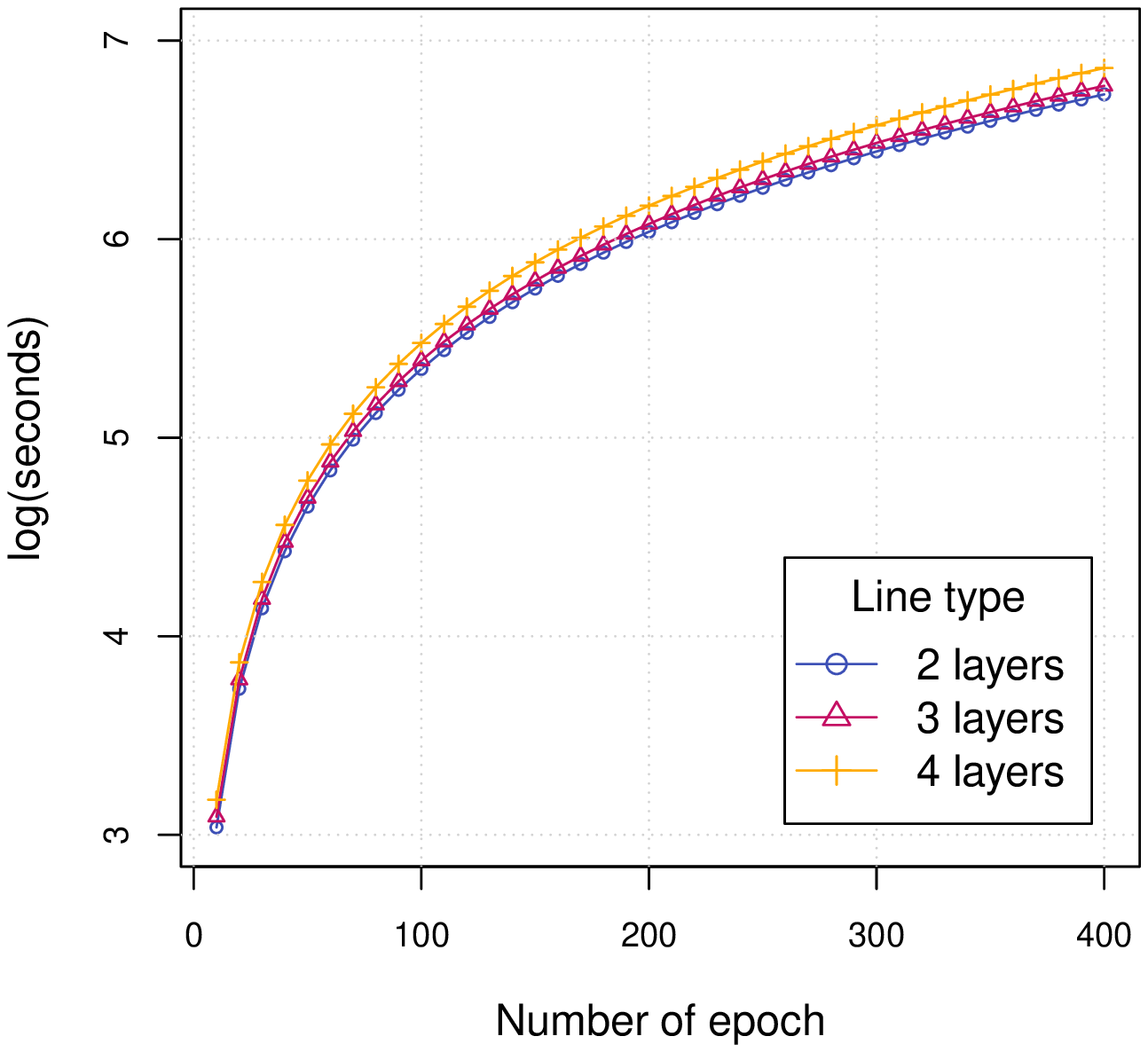}
\centering{(i)}
\end{minipage}
\begin{minipage}[t]{6.15cm}
\centering
\includegraphics[width=5.6cm]{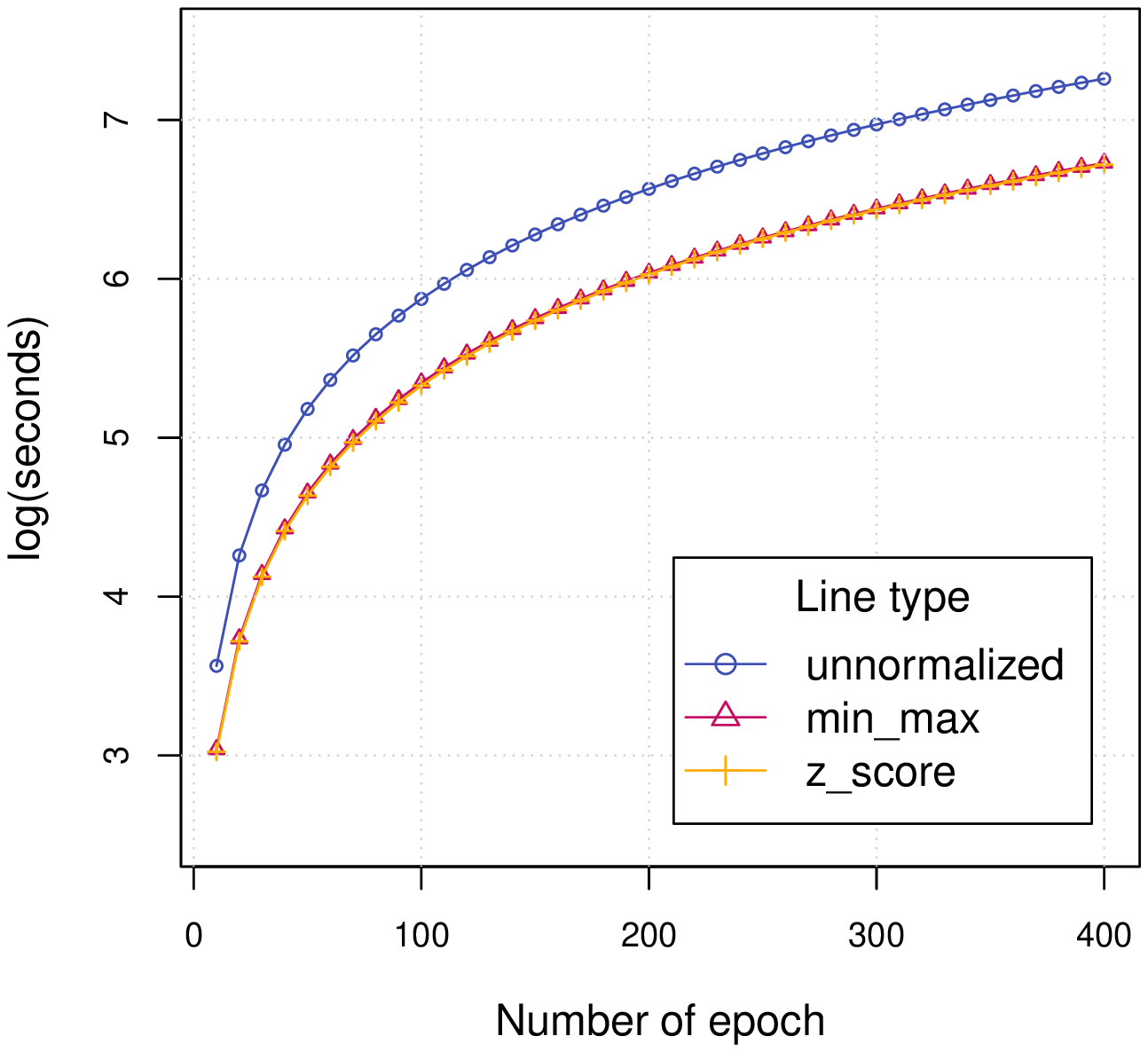}
\centering{(j)}
\end{minipage}
\begin{minipage}[t]{6.15cm}
\centering
\includegraphics[width=5.6cm]{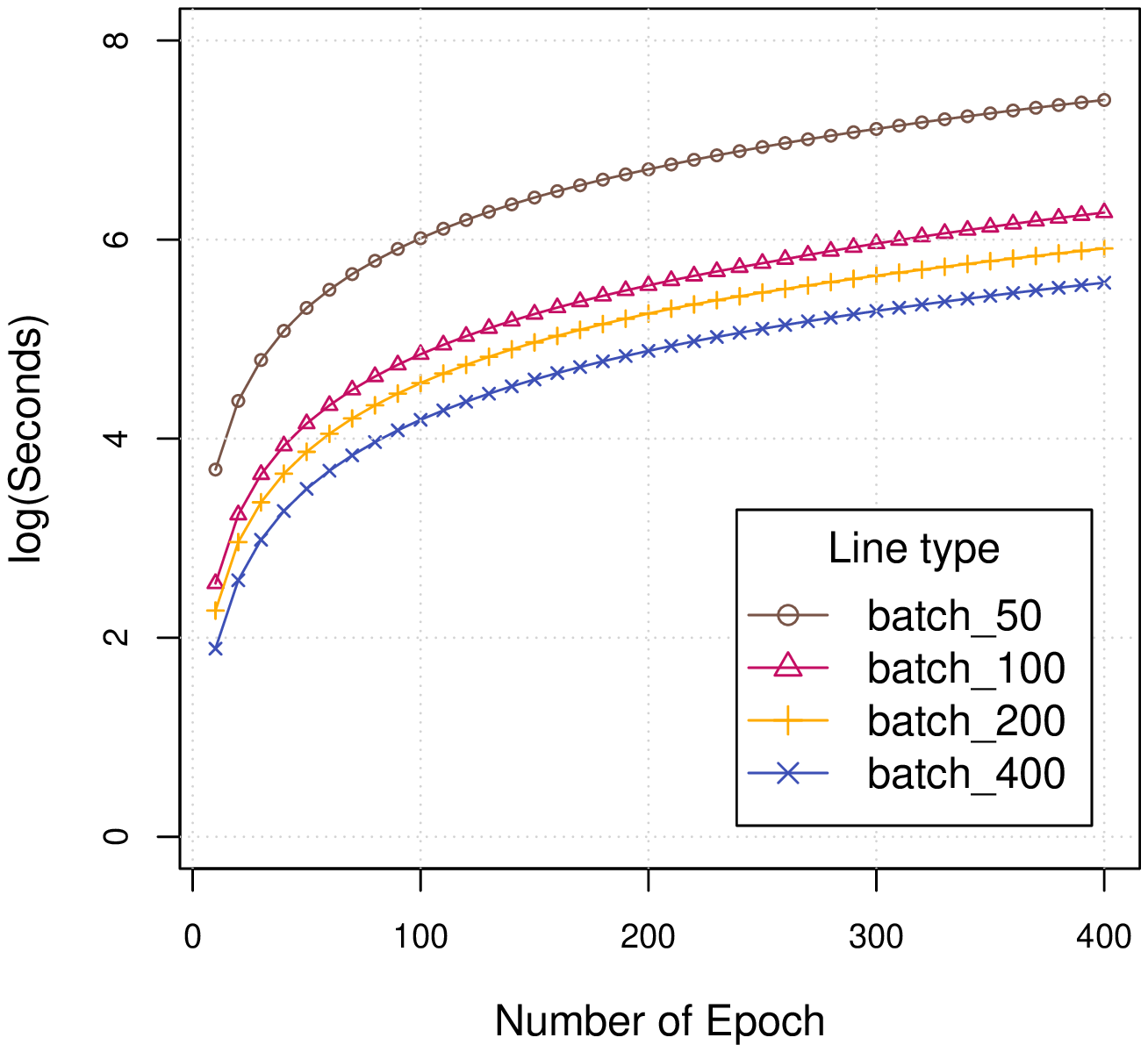}
\centering{(k)}
\end{minipage}
\begin{minipage}[t]{6.15cm}
\centering
\includegraphics[width=5.6cm]{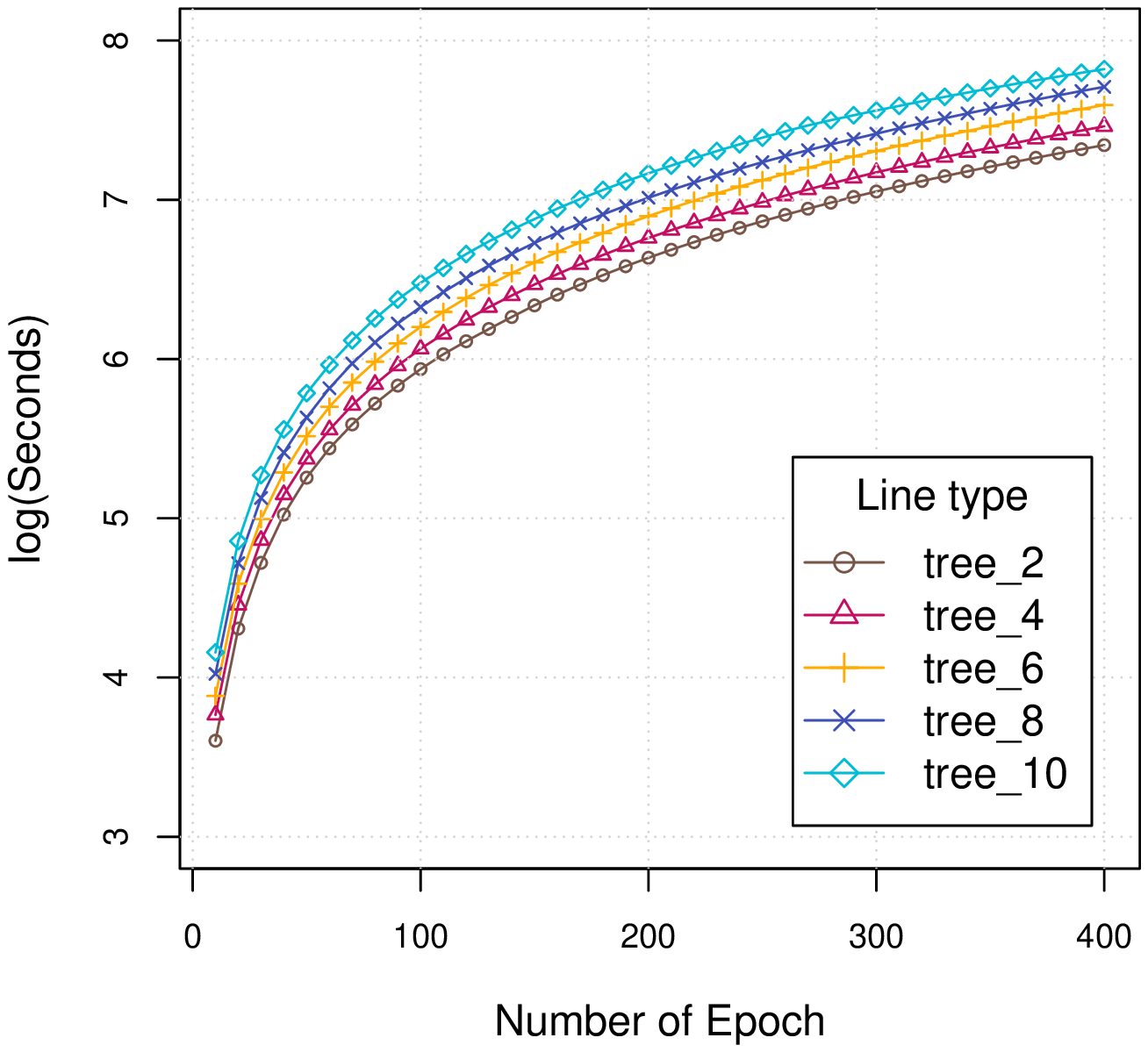}
\centering{(l)}
\end{minipage}
\caption{The accuracy for 400 epochs of the parameter: (a)the number of depths of the tree, (b)with or without fully connected layers between autoencoder and neural decision forest, (c)the number of autoencoder layers, (d)the normalization methods, (e) the batch size, and (f) the number of trees in a forest. And the time for 400 epoches of the according parameters: (g)-(l).}
\label{fig:results}
\end{figure*}

We compared the accuracy and time cost among different parameters, which is shown in figure \ref{fig:results}.

From Figure~\ref{fig:results} (a)-(f) we can see that the
the random forest with only 1 depth outputs a bad prediction in performance, and the accuracy becomes better with 2 and onwards in depths; fully connected layers can help with the prediction accuracy; a model with 2 layers in autoencoder can get great prediction accuracy with first 100 epochs, and more layers for each autoencoder results in better prediction yet longer to converge and give the result; the normalized dataset shows better performance than the unnormalized one, and z\_score has a better performance than min-max in terms of normalization; batch size 50 produces the optimal prediction, with a comparable short computation time at the highest prediction accuracy; and finally, the prediction improves with increased number of trees in a forest.


Figure~\ref{fig:results} (g)-(l) show the time consumption under different parameter settings. The axis y is taking the $log$ for the time (seconds). We could see deeper the depth of the tree, more time will be cost; add fully connected layers will increase the time cost; the time cost will increase with bigger number of autoencoder layers; unnormalized dataset cost more time; less batch size takes more training time; and higher tree depths incurs more training time.

Thus, considering the time cost and the prediction performance, we choose a batch size with 50, 5 trees in a random forest with depth of 3, z-score as normalization method, one fully connected layer, and 2 layers in autoencoder. In the test we get correctly predicted fake reviews and 2192 correctly predicted genuine reviews, 65 falsely predicted fake reviews and 99 falsely predicted genuine reviews. This means an accuracy of approximately 95.85\%.

\subsection{Overall Comparison}
We compared our model with a set of existing works, which also use Amazon dataset but are different in extracted features and solutions. A brief introduction for these papers is listed below:

\begin{itemize}
  \item Jindal and Liu \citep{jindal2008opinion}：first manually labeled easy identify spam reviews and then considered multi-features, they used logistic regression to do the classification.
  \item Lai et al. \citep{lai2010toward}: considered the content of the reviews, they used an unsupervised probabilistic language model and a supervised SVM model.
  \item Mukherjee et al. \citep{mukherjee2013spotting}: This work used unsupervised Bayesian approach and considered user's behavior features and bigrams.
  \item Rout et al. \citep{rout2017deceptive}: considered the sentiment analysis and they extracted sentiment score, linguistic features and unigram as feature set.
  \item Shreesh et al.\citep{bhat2017identifying}: mainly focused on the content based information and they combined positive unlabeled learning with domain adaptation to train a classifier.
  \item Heydari et al. \citep{heydari2016detection}: Heydari et al. considered the rating deviation, content based factors and activeness of reviewers, and they captured suspicous time intervals from time series by a pattern recognition technique.
  \item Xu and Zhang \citep{xu2015towards}:  considered target based, rating based, temporal based and activity based signals. A Latent Collusion Model (LCM) is proposed to work unsupervisedly and supervisedly under different conditions.
  \item Zhang et al. \citep{zhang2016online}: considered unverbal features and used four classical supervised classification methods: SVM, Naive Bayes, decision tree and random forest.
\end{itemize}

To evaluate the performance of our fake news detection algorithms, we used the following criteria \citep{powers2011evaluation}:

\begin{equation}
\begin{split}
Accuracy =& \frac{TP+TN}{TP+FP+TN+FN} \\
Precision =& \frac{TP}{TP+FP} \\
Recall =& \frac{TP}{TP+FN} \\
F1 =& \frac{2TP}{2TP+FP+FN}
\end{split}
\end{equation}
where TP, FP, TN, FN represent the counts for true positive, false positive, true negative and false negative, respectively.

Table 3 shows the comparison results. And the results for each paper are extracted from the published papers. We could see that our proposed model gives the best prediction accuracy, precision and F1 score values.

Comparing with these methods, we could conclude that our approach demonstrates readily good performance in predicting the fake reviews. It also indicates that the proposed neural autoencoder random forest model is at least a appreciable classification model in predicting the review spam.

\begin{table}[ht]
\caption{Comparison with the State-of-art Methods}
\label{tab:finalresult}
\small
\centering
\begin{tabular}{l|l|l|l|l}
  \hline
  \textbf{Methods} & \textbf{Precision} & \textbf{Accuracy} & \textbf{F1} & \textbf{Recall}\\ \hline
  Jindal and Liu & - & 78\% & - & -\\
  Lai et al. & - & - & - & \textbf{96.38\%} \\
  Mukherjee et al. & 79.6\% & 86.1\% & 77.3\% & 75.1\%\\
  Rout et al. & - & 92.11\% & - & -\\
  Shreesh et al. & 87.6\% & - & 84.3\% & 82.8\%   \\
  Heydari et al. & 82\% & - & 86\% & 88\% \\
  Xu and Zhang & 86.4\% & 85.2\% & - & - \\
  Zhang et al. & 87.12\% & 87.81\% & 88.31\% & 89.63\% \\  \hline
  Ours & \textbf{96.08\%} & \textbf{95.85\%} & \textbf{95.11\%} & 94.15\% \\
  \hline
\end{tabular}
\label{tab:states}
\end{table}

\section{Conclusion}
In this paper, we focus on detecting fake reviews, which is an attractive topic and has drawn attention of certain researchers. While previous work mainly focused on using classical classification methods, here we proposed an end to end trainable joint model combining autoencoder with neural random forest, to detect the fake reviews. Firstly, we did quality feature analysis to mining the feature efficiency in detecting opinion fraud, and those efficient features are employed as input in our model. And then we declared the design of our model. The model starts with an autoencoder, and then fully connected layers and ends in a differentiable random forest, where the differentiable random forest is trained back propagation. The forest averages the prediction result of each tree and outputs our final prediction. We also analyzed the impact of features and tuned various parameters in our model. The extensive experimental results demonstrate that our method beats a series of state-of-the-art methods yielding 96\% of accuracy.

\bibliographystyle{model2-names}
\bibliography{refs}

\end{document}